\documentclass[nohyperref]{article}

\usepackage{microtype}
\usepackage{graphicx}
\usepackage{subfigure}
\usepackage{booktabs} 

\usepackage{hyperref}

\usepackage[accepted]{icml2022}

\usepackage{amsmath}
\usepackage{amssymb}
\usepackage{mathtools}
\usepackage{amsthm}

\usepackage{natbib}
\usepackage{enumitem}
\usepackage{multirow}
\usepackage{bm}
\usepackage{array}

\newcommand{\ie}{\textit{i}.\textit{e}.,}
\newcommand{\eg}{\textit{e}.\textit{g}.,}
\newcommand{\etc}{\textit{etc}.}
\newcommand{\indep}{\perp \!\!\! \perp}

\usepackage[capitalize,noabbrev]{cleveref}

\theoremstyle{plain}
\newtheorem{theorem}{Theorem}[section]

\theoremstyle{definition}

\theoremstyle{remark}

\usepackage[textsize=tiny]{todonotes}

\icmltitlerunning{Generalizing to Evolving Domains with Latent Structure-Aware Sequential Autoencoder}

\begin{document}

\twocolumn[
\icmltitle{Generalizing to Evolving Domains with Latent Structure-Aware \\ Sequential Autoencoder}



\icmlsetsymbol{equal}{*}

\begin{icmlauthorlist}
\icmlauthor{Tiexin Qin}{yyy}
\icmlauthor{Shiqi Wang}{yyy}
\icmlauthor{Haoliang Li}{yyy}
\end{icmlauthorlist}

\icmlaffiliation{yyy}{City University of Hong Kong, Hong Kong}

\icmlcorrespondingauthor{Haoliang Li}{haoliang.li1991@gmail.com}

\icmlkeywords{Machine Learning, Domain Generalization, Variational Inference}

\vskip 0.3in
]



\printAffiliationsAndNotice{}  

\begin{abstract}
Domain generalization aims to improve the generalization capability of machine learning systems to out-of-distribution (OOD) data. Existing domain generalization techniques embark upon stationary and discrete environments to tackle the generalization issue caused by OOD data. However, many real-world tasks in non-stationary environments ~(\eg~self-driven car system, sensor measures) involve more complex and continuously evolving domain drift, which raises new challenges for the problem of domain generalization. In this paper, we formulate the aforementioned setting as the problem of \emph{evolving domain generalization}. Specifically, we propose to introduce a probabilistic framework called Latent Structure-aware Sequential Autoencoder~(LSSAE) to tackle the problem of evolving domain generalization via exploring the underlying continuous structure in the latent space of deep neural networks, where we aim to identify two major factors namely \emph{covariate shift} and \emph{concept shift} accounting for distribution shift in non-stationary environments. Experimental results on both synthetic and real-world datasets show that LSSAE can lead to superior performances based on the evolving domain generalization setting. 
\end{abstract}

\section{Introduction}
\label{sec:intro}
The success of machine learning techniques typically lies on the assumption that training data and test data are sampled independently and identically from similar distributions. However, this assumption does not hold when deploying the trained model in many real-world environments where the distribution of test data varies from training data. This distribution discrepancy, so-called distribution shift, can lead to the dramatic performance decrease of machine learning models~\cite{Torralba2011CVPR}. To mitigate this issue, domain generalization (DG) has been proposed to learn a more robust model which can be better generalized to OOD data~\cite{Muandet2013ICML, Balaji2018NIPS, Li2018MMD}. 


While some progress is being achieved so far, existing DG methods are limited to the setting of generalization among discrete and stationary environments. This setting can be problematic in some real-world applications where we require that model can be generalized among continuous domains~\cite{Hoffman2014CVPR}. For example, a self-driving car system, when deployed in the real world, struggles to perform under an open environment where the accepted data changes naturally according to the geographic location, time intervals, and other factors in a gradual manner~\cite{Hoffman2014CVPR}. For another example, the measures of sensors can also drift over time due to the outer environments and inter factors, such as aging~\cite{Vergara2012Chemical}. In these scenarios, treating each domain in a separate manner is unlikely to yield the desired performance as it does not consider the property of continuous domain structure. 


Another limitation of most of the existing DG methods is that they did not take ``concept shift" into consideration. Such concept shift can also lead to performance drop~\cite{Federici2021Theory}.  A typical example of concept shift would be that the incidence rate of a particular disease in certain groups may change over time due to the development of treatments and preventive measures. Therefore, existing DG techniques may fail to be applied to some other complex real-world applications in non-stationary environments~\cite{Sugiyama2013WIRCS,Tahmasbi2021DriftSurf}.

In this paper, we propose to focus on the problem of domain generalization based on the non-stationary setting, where data can evolve gradually with both covariate shift and concept shift. Particularly, we formulate this non-stationary scenario as \emph{evolving domain generalization} where we only have access to adequate labeled examples from the sequential source domains. Our objective is to develop algorithms that can explore the underlying continuous structure of distribution shift and generalize well to evolving target domains where the samples are unavailable during the training stage. Unlike existing DG methods that only focus on covariate shift based on the stationary environment, in this paper, we propose a novel framework called Latent Structure-aware Sequential Autoencoder~(LSSAE), a dynamic probabilistic framework to model the underlying latent variables across domains. More specifically, we propose to use two latent variables to represent the sampling bias in data sample space (i.e., covariate shift) and data category space (i.e., concept shift), and propose a domain-related module and a category-related module to infer their dynamic transition functions based on different time stamps. We conduct extensive experiments to verify that our framework can successfully capture the underlying covariate shift and interpret the concept shift simultaneously. Last but not least, we show that our proposed LSSAE has a promising generation capability to predict unseen target domains, which can be helpful to the problem related to sequential data generation. The main contributions of this paper are summarized as follows.
\begin{itemize}[noitemsep]

\item We propose to focus on the problem of non-stationary evolving domain generalization where both covariate shift and concept shift may exist in the setting.

\item We propose a novel probabilistic framework LSSAE which incorporates variational inference to identify the continuous latent structures of these two shifts separately and simultaneously.

\item We provide empirical results to show that the proposed approach yield better results than other DG methods across scenarios. Besides, it presents a  powerful generation ability of predicting unseen evolving domains.
\end{itemize}

\section{Related Work}
\label{sec:related_work}

\noindent{\textbf{Domain Generalization (DG)}.}
The goal of DG is to learn robust models which can generalize well towards the out-of-distribution samples from unseen domains. Existing DG methods commonly rely on multiple source domains to learn representative features that can be better generalized. According to various strategies used to learn these representations, we can roughly categorize them into three catogories. The first type is feature-based methods, which aim to learn domain-invariant representation which can be better generalized to target domains. Specifically, it can be achieved by aligning the distribution of representations from all source domains~\cite{Blanchard2011NIPS,Li2018MMD,Albuquerque2021arXiv} and feature disentanglement~\cite{Ilse2020Diva,Wang2021arXiv,Nguyen2021NIPS}. The second category is meta-learning based methods, which utilize the model agnostic training procedure to stimulate the train/test shift for acquiring generalized models~\cite{Li2018AAAI,Balaji2018NIPS,Dou2019NIPS}. Last but not least, data augmentation-based techniques, which aim to manipulate the perturbation both in original images and features to stimulate the unseen target domains, can also benefit the problem of domain generalization~\cite{Volpi2018NIPS,Shankar2018ICLR,Zhou2021arXiv}. 

\noindent{\textbf{Continuous Domain Adaptation (CDA).}}
The problem of continuous domain adaptation (i.e., evolving domain adaptation) has attracted increasing attention recently, where the CDA methods can be categorized into the intermediate-domains based methods~\cite{Kumar2020ICML,Gong2019CVPR,Chen2021NIPS}, domain manifold based methods~\cite{Hoffman2014CVPR,Li2017PAMI}, adversary-based approaches~\cite{Wang2020CIDA,Wulfmeier2018ICRA}, and meta-learning based methods~\cite{Long2020EDA,Lao2020Continuous}. More or less, they require some samples from target domains for adaptation. In~\citet{Mancini2019CVPR}, they propose to substitute this reliance with some metadata from target domains as additional supervision. Instead, our focus is continuous domain generalization where no information from target domains is accessible for model learning, which is a more challenging but realistic task for real-world applications.

\noindent{\textbf{Sequential Data Generation.}}
Recent process in unsupervised sequence generation~\cite{Li2018DSAE,Han2021RWAE,Park2021ICML} suggests the importance of decoupling time-invariant and time-variant information during the representation learning procedure. However, these approaches only take sequence generation tasks into consideration and fail to consider the category-related information, which is important for the problem of domain generalization. Unlike these approaches, we propose jointly focusing on the dynamic modeling of time-variant information on both data sample space and category space across domains. 


\section{Methodology}
\label{sec:method}

In this section, we first formalize the problem of evolving domain generalization based on the non-stationary environment and then describe our framework LSSAE for addressing this problem. After that, we will provide theoretical analysis for the proposed framework in Sec.~\ref{sec:theoretical_analysis} and implementation in Sec.~\ref{sec:imp}.

\subsection{Problem Formulation}

\begin{figure}[!tbp]
\begin{center}
\includegraphics[width=0.48\textwidth]{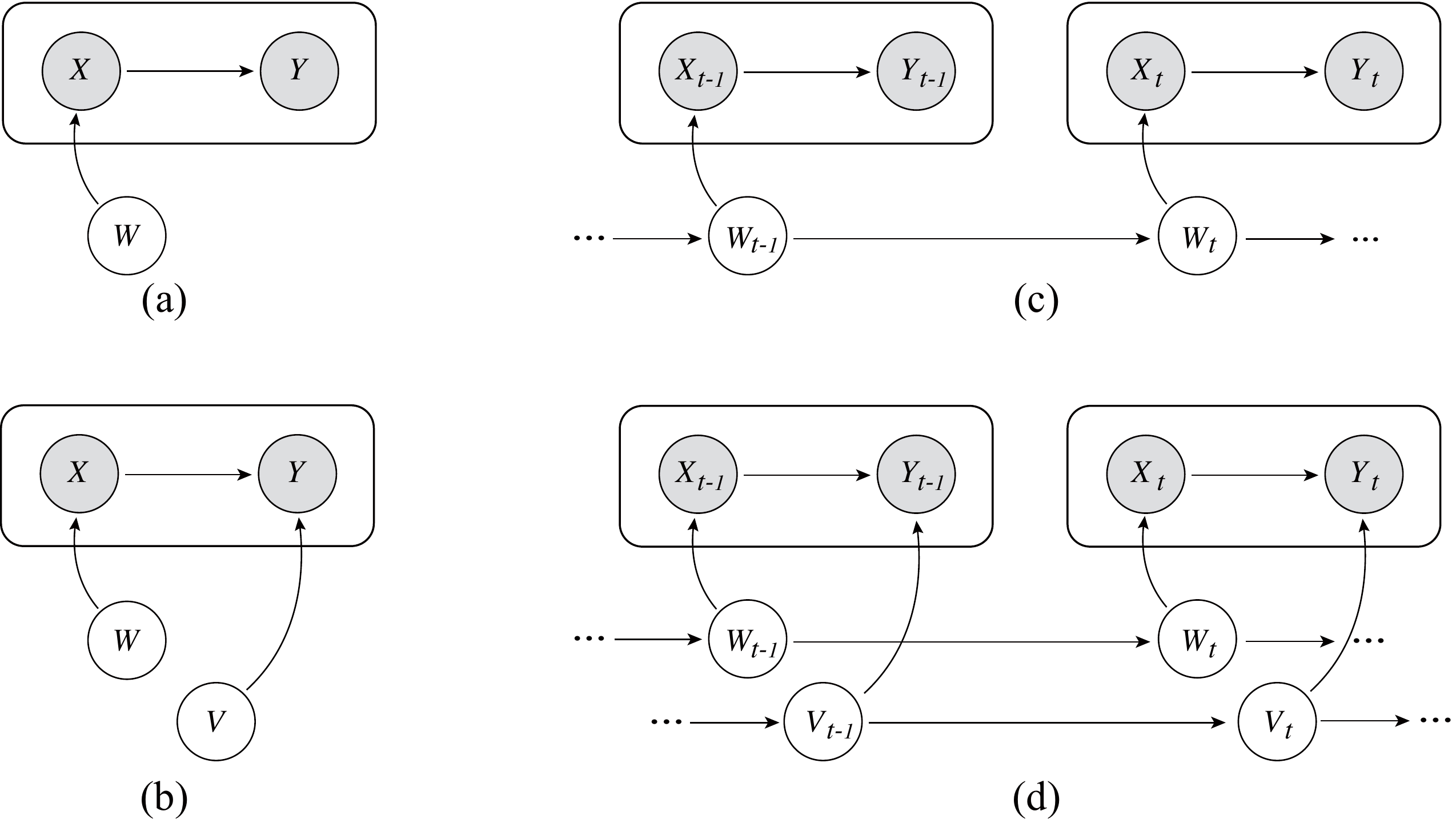}
\end{center}
\vspace{-0.2cm}
\caption{Comparison of causality diagram for stationary and non-stationary domain generalization scenarios. (a) represents the standard stationary DG settings which only contains covariate shift ($P(X)$ varies for source and target domains). (b) is an extension version of (a) which contains both covariate shift and concept shift $(P(Y|X)$ varies). (c) and (d) are corresponding non-stationary DG settings where there exist evolving patterns among adjacent domains.} 
\label{fig:DAG}
\end{figure}

Suppose we are given $T$ sequentially arriving \emph{source} domains $\mathcal{S}=\{ \mathcal{D}_1,\mathcal{D}_2,...,\mathcal{D}_T \}$, where each domain $\mathcal{D}_t=\{{(x_{t,i},y_{t,i})\}_{i=1}^{n_{t}}}$ is comprised of $n_{t}$ labeled samples for $t \in \{1,2,...,T\}$. The goal of our problem setting is to train a classification model on $\mathcal{S}$ which can be generalized to $M$ following arriving \emph{target} domains $\mathcal{T}=\{ \mathcal{D}_{T+1},\mathcal{D}_{T+2},...,\mathcal{D}_{T+M}\}$, $\mathcal{D}_{t}=\{{(x_{t,i})\}_{i=1}^{n_{t}}}$ ($t \in \{T+1,T+2,...,T+M\}$), which are not available during training stage. For simplicity, we omit the index $i$ whenever $\mathbf{x}_{i}$ refers to a single data point.  To further quantify the continuously evolving nature of domains, we denote $0 \leq \mathbb{D}(\mathcal{D}_{t}, \mathcal{D}_{t+1}) \leq \epsilon$ for two consecutive domains under some distribution distance function $\mathbb{D}$ (e.g., Kullback-Leibler distance). In other words, the discrepancy between two consecutive domains is bounded.


Conventional DG setting only assumes that $P(X)$ varies~(i.e., covariate shift) for different domains~(See Fig.~\ref{fig:DAG}~(a)), which may not be ideal since both  $P(X)$ and  $P(Y|X)$ can be non-stationary (i.e., $P(X)$ and $P(Y|X)$ vary over time which lead to evolving covariate shift and concept shift, respectively). To tackle this problem, in this paper, we aim to explore the evolving patterns of covariate shift and concept shift across domains.



\subsection{LSSAE:~Latent Structure-aware Sequential Autoencoder}
\label{sec:LSSAE}

To model the dynamic in non-stationary systems, we consider two independent factors $W$ and $V$~(i.e., $W \indep V$) which account for the distribution drift in data sample space (i.e., covariate shift) and data category space (i.e., concept shift) respectively according to different time stamps (See~Fig.~\ref{fig:DAG}~(d)). For data $(\mathbf{x}_t, \mathbf{y}_t)$ collected at time stamp $t$, we denote $\mathbf{z}_t^w$ and $\mathbf{z}_t^v$ as the latent variables of $W$ and $V$ at time stamp $t$.  For completeness, we further consider  time-invariant latent code $\mathbf{z}^c$ to capture the static information of $\mathbf{x}_t$. We thus can define a probabilistic generative model for the joint distribution of all source domains as
\begin{equation}\label{equ:joint_xy}
\begin{split}
p&(\mathbf{x}_{1:T},\mathbf{y}_{1:T},\mathbf{z}^c,\mathbf{z}_{1:T}^w, \mathbf{z}_{1:T}^v) \\
&= p(\mathbf{x}_{1:T}, \mathbf{z}_{1:T}^w, \mathbf{z}^c) p(\mathbf{y}_{1:T}, \mathbf{z}_{1:T}^v| \mathbf{z}^c).
\end{split}
\end{equation}
where the first term $p(\mathbf{x}_{1:T}, \mathbf{z}_{1:T}^w, \mathbf{z}^c)$ and second term $p(\mathbf{y}_{1:T}, \mathbf{z}_{1:T}^v| \mathbf{x}_{1:T})$ can be formulated by using Markov chain model as
\begin{equation}\label{equ:gen_x}
    p(\mathbf{x}_{1:T}, \mathbf{z}_{1:T}^w, \mathbf{z}^c) = p(\mathbf{z}^c) \!\displaystyle\prod_{t=1}^T\!
     p(\mathbf{z}_{t}^w|\mathbf{z}_{<t}^w) \underbrace{p(\mathbf{x}_{t}|\mathbf{z}^c,\mathbf{z}_{t}^w)}_{covariate~shift},
\end{equation}
\begin{equation}
\label{equ:gen_y}
    p(\mathbf{y}_{1:T}, \mathbf{z}_{1:T}^v| \mathbf{z}^c) = \!\displaystyle\prod_{t=1}^T\!
    p(\mathbf{z}_{t}^v|\mathbf{z}_{<t}^v) \underbrace{p(\mathbf{y}_{t}|\mathbf{z}^c,\mathbf{z}_{t}^v)}_{concept~shift},
\end{equation}
and $p(\mathbf{x}_{t}|\mathbf{z}^c,\mathbf{z}_{t}^w)$ and $p(\mathbf{y}_{t}|\mathbf{z}^c,\mathbf{z}_{t}^v)$ denote covariate shift and concept shift, respectively. 
 Eq.~\ref{equ:gen_x} shows that the generation process of  domains data $\mathbf{x}_t$ at time stamp $t$ depends on the corresponding dynamic latent code $\mathbf{z}_t^w$ and static code $\mathbf{z}^c$, and Eq.~\ref{equ:gen_y} shows that the inference process (i.e.,  classifier to produce $\mathbf{y}_t$) rely on the corresponding $\mathbf{z}_t^v$ and $\mathbf{z}^c$. Our objective is to learn the  classifier $p(\mathbf{y}_{t}|\mathbf{z}^c,\mathbf{z}_{t}^v)$ which disposes of covariate shift through $\mathbf{z}^c$ and concept shift with dynamic $\mathbf{z}_{t}^v$ for the problem of evolving domain generalization.



\noindent\textbf{Domain-related module for covariate shift.}~To model $p(\mathbf{x}_{1:T}, \mathbf{z}_{1:T}^w, \mathbf{z}^c)$ where covariate shift involved, we set the prior distribution as $p(\mathbf{z}^c)=\mathcal{N}(\mathbf{0},\mathbf{I})$, $p(\mathbf{z}_{t}^w|\mathbf{z}_{<t}^w)=\mathcal{N}(\bm{\mu}(\mathbf{z}^w_t),\bm{\sigma}^2(\mathbf{z}_{t}^w))$ which can be parameterized by some recurrent neural networks (\eg~LSTM~\cite{Hochreiter1997LSTM}) by setting $\mathbf{z}_{0}^w=\mathbf{0}$ for initial state when $t=0$, and $p(\mathbf{x}_{t}|\mathbf{z}^c,\mathbf{z}_{t}^w)$ as a conditional decoder for the reconstruction of input data $\mathbf{x}_{t}$. 
To approximate the prior distributions $p(\mathbf{z}_{t}^w|\mathbf{z}_{<t}^w)$, we propose to use variational inference to learn an approximate posterior distribution $q$ over latent variables given data which can be formulated as
\begin{equation}
    q(\mathbf{z}_{1:T}^w,\mathbf{z}^c|\mathbf{x}_{1:T}) =
    q(\mathbf{z}^c|\mathbf{x}_{1:T}) \displaystyle\prod_{t=1}^T
    q(\mathbf{z}_t^w|\mathbf{z}_{<t}^w,\mathbf{x}_t),
\end{equation}
where $q(\mathbf{z}^c|\mathbf{x}_{1:T})$ and $q(\mathbf{z}_t^w|\mathbf{z}_{<t}^w,\mathbf{x}_t)$ can be also parameterized by neural networks. The objective function for latent feature representation learning can be derived based on the evidence lower bound (ELBO) form \cite{kingma2014VAE} given as
\begin{equation}
\label{equ:loss_d}    
\begin{split}
    \mathcal{L}_d &= \displaystyle\sum_{t=1}^T\mathbb{E}_{q(\mathbf{z}^c|\mathbf{x}_{t}) q(\mathbf{z}_t^w|\mathbf{z}_{<t}^w,\mathbf{x}_t)} \big[\log p(\mathbf{x}_{t}|\mathbf{z}^c,\mathbf{z}_{t}^w) \big] \\
    &- \lambda_{1} \mathbb{D}_{KL}(q(\mathbf{z}^c|\mathbf{x}_{1:T}),p(\mathbf{z}^c)) \\
    &- \lambda_{2} \displaystyle\sum_{t=1}^T \mathbb{D}_{KL} (q(\mathbf{z}_t^w|\mathbf{z}_{<t}^w,\mathbf{x}_t),p(\mathbf{z}_{t}^w|\mathbf{z}_{<t}^w)),
\end{split}
\end{equation}
where the first term denotes the reconstruction term for input data $\mathbf{x}_t$, the second and third term denote KL divergence which are to align the posterior distributions $\mathbf{z}^c$ and $\mathbf{z}^w_t$ with the corresponding prior distributions. 




\noindent\textbf{Category-related module for concept shift. }~To model $p(\mathbf{y}_{1:T}, \mathbf{z}_{1:T}^v| \mathbf{z}^c)$ for classification purpose where concept shift is involved, we propose to introduce another module which can be easily integrated into our unified framework under domain generalization scenarios. Specifically, we propose to model $p(\mathbf{y}_{1:T}, \mathbf{z}_{1:T}^v| \mathbf{z}^c)$ with a dynamic distribution $p(\mathbf{y}_{t},\mathbf{z}_{t}^v|\mathbf{z}^c)$  where the  $\mathbf{z}^v_t$ is varied and encode the shift information in the data category space (\eg~the proportion of each category). The module can be optimized via maximizing the distribution $p(\mathbf{y}_{1:T}, \mathbf{z}_{1:T}^v| \mathbf{z}^c)$ given a sequence of domains.
In practise, we represent $p(\mathbf{z}_{t}^v|\mathbf{z}_{<t}^v)$ as $p(\mathbf{z}_{t}^v|\mathbf{z}_{<t}^v)=\text{Cat}(\bm{\pi}(\mathbf{z}_{<t}^v))$  which is a learnable categorical distribution. In a similar vein with $\mathbf{z}_t^w$, we utilize a distribution $q$ to model the posterior distribution and approximate the prior distribution of $\mathbf{z}_t^v$ by adopting variational inference given as
\begin{equation}
\label{equ:infer_v}
    q(\mathbf{z}_{1:T}^v|\mathbf{y}_{1:T}) = \displaystyle\prod_{t=1}^T 
    q(\mathbf{z}_t^v|\mathbf{z}_{<t}^v, \mathbf{y}_t),
\end{equation}
where $q(\mathbf{z}_t^v|\mathbf{z}_{<t}^v,\mathbf{y}_t)$ can be parameterized by a recurrent neural network with categorical distribution as output. We set $\mathbf{z}_{0}^v=\mathbf{0}$ for initial state when $t=0$. The proposed module can be jointly trained with the inference process in Eq.~\ref{equ:infer_v} as well as classification loss by maximizing
\begin{equation}
\label{equ:loss_c}    
\begin{split}
    \mathcal{L}_c &= \displaystyle\sum_{t=1}^T\mathbb{E}_{q(\mathbf{z}_t^v|\mathbf{z}_{<t}^v,\mathbf{y}_t)} \big[\log p(\mathbf{y}_{t}|\mathbf{z}^c,\mathbf{z}_{t}^v) \big] \\
    &- \lambda_{3} \displaystyle\sum_{t=1}^T \mathbb{D}_{KL} (q(\mathbf{z}_t^v|\mathbf{z}_{<t}^v,\mathbf{y}_t),p(\mathbf{z}_{t}^v|\mathbf{z}_{<t}^v)).
\end{split}
\end{equation}
Here, the first term denotes the classification loss (i.e., maximizing log likelihood) and the second term denotes KL divergence which aims to align the posterior distribution of $\mathbf{z}^v_t$ with its prior distribution.




\noindent\textbf{Temporal smooth constraint for better stability.}~ We empirically find that the estimation of conditional density~(i.e.,~ $p(\mathbf{z}_{t}^w|\mathbf{z}_{<t}^w)$ and $p(\mathbf{z}_{t}^v|\mathbf{z}_{<t}^v)$) which is to model the complex dynamics over temporal transition may not be stable based on our formulation above. We conjecture the reason that some of the static information can be distorted by the dynamic inference module $q(\mathbf{z}_t^w|\mathbf{z}_{<t}^w,\mathbf{x}_t)$ and $q(\mathbf{z}_t^v|\mathbf{z}_{<t}^v,\mathbf{y}_t)$ for better reconstruction quality, which further yields sub-optimal results for recognition tasks. Intuitively, one can tackle this limitation by reducing the dimension of latent codes $\mathbf{z}^w_t$ and $\mathbf{z}^v_t$ or decreasing the learning rate of corresponding inference modules and prior modules manually to achieve better decoupling effect. However, we find that such strategy may not lead to desired performance. In our work, we propose to employ Lipschitz constrain over the temporal domain to stabilize the learning of the dynamic inference modules as follows
\begin{equation}
\label{equ:ts}
\begin{split}
    |q(\mathbf{z}_t^w|\mathbf{z}_{<t}^w, \mathbf{x}_t) - q(\mathbf{z}_{t-1}^w|\mathbf{z}_{<t-1}^w, \mathbf{x}_{t-1})| \leq \alpha, \\
    |q(\mathbf{z}_t^v|\mathbf{z}_{<t}^v, \mathbf{y}_t) - q(\mathbf{z}_{t-1}^v|\mathbf{z}_{<t-1}^v, \mathbf{y}_{t-1})| \leq \alpha,
\end{split}
\end{equation}
where $\alpha$ is referred to as a Lipschitz constant. The above regularization term is denoted as TS constraint for simplicity. We expect it can help with reducing the potential for volatile training.



\noindent\textbf{Objective function.}~Given training data $\mathcal{S}$, our proposed framework can be optimized through the objective function $\mathcal{L}_{LSSAE}=\mathcal{L}_d + \mathcal{L}_c$ with the temporal smooth constrains in Eq.~\ref{equ:ts}, where the first term $\mathcal{L}_d$ and second term $\mathcal{L}_c$ aim to tackle the problem of covariate shift and concept shift, respectively. 


\textbf{Discussion. } It is worth noting that there exists some works using probabilistic graph model for future video frame generation task~\cite{Li2018DSAE, Han2021RWAE}, which are similar to our proposed method at a high level. Nevertheless, our method is different since 1) our proposed framework can be applied not only to generation task (as we show in the ablation study of experimental section) but also to evolving domain generalization task (which is the main focus in our paper); 2) in order to fit our framework to the non-stationary recognition task, we introduce a novel category-related module to capture the concept shift, as such, better generalization performance can be achieved. 


\subsection{Theoretical Analysis}
\label{sec:theoretical_analysis}
In this section, we aim to give a theoretical insight on our proposed method by extending variational inference from stationary environments to non-stationary environments.

\noindent\textbf{Probabilistic model for stationary environments.}~We first elaborate our proposed framework based on the stationary condition, where $\mathbf{z}^c$, $\mathbf{z}^w$ and $\mathbf{z}^v$ are introduced to capture the domain-invariant category information, domain-specific and category information, respectively. We thus have the following theorem.


\begin{theorem}
\label{theorem:static}
For the data log likelihood $\log p(\mathbf{x}, \mathbf{y})$ in a stationary environment, we have the evidence lower bound
\begin{equation}
\label{equ:static_elbo}
    \begin{aligned}
        \displaystyle \max_{p,q}~& \mathbb{E}_{\mathbf{z}^c,\mathbf{z}^w,\mathbf{z}^v} \big[ \log p(\mathbf{x}|\mathbf{z}^c,\mathbf{z}^w)p(\mathbf{y}|\mathbf{z}^c,\mathbf{z}^v) \big] \\
        &- \mathbb{D}_{KL}(q(\mathbf{z}^c|\mathbf{x}),p(\mathbf{z}^c)) - \mathbb{D}_{KL}(q(\mathbf{z}^w|\mathbf{x}),p(\mathbf{z}^w)) \\
        &- \mathbb{D}_{KL}(q(\mathbf{z}^v|\mathbf{y}),p(\mathbf{z}^v)),
    \end{aligned}
\end{equation}
where $\mathbf{z}^c \sim q(\mathbf{z}^c|\mathbf{x}), \mathbf{z}^w \sim q(\mathbf{z}^w|\mathbf{x}), \mathbf{z}^v \sim q(\mathbf{z}^v|\mathbf{y})$.
\end{theorem}

\emph{Proof.}~We consider the data generation procedure $p(\mathbf{x}, \mathbf{y}, \mathbf{z}^c, \mathbf{z}^w, \mathbf{z}^v)$. We have
\begin{equation}
\label{equ:static_log}
    \begin{aligned}
        \log p(\mathbf{x}, \mathbf{y}) &= \mathbb{D}_{KL}(q(\mathbf{z}^c, \mathbf{z}^w, \mathbf{z}^v |\mathbf{x}, \mathbf{y}), p(\mathbf{z}^c, \mathbf{z}^w, \mathbf{z}^v |\mathbf{x}, \mathbf{y})) \\
        &+ \mathbb{E}_q \log \frac{p(\mathbf{x}, \mathbf{y}, \mathbf{z}^c, \mathbf{z}^w, \mathbf{z}^v)}{q (\mathbf{z}^c, \mathbf{z}^w, \mathbf{z}^v | \mathbf{x}, \mathbf{y})},
    \end{aligned}
\end{equation}
where the first term is the KL divergence of the approximate from the true posterior. Since this term is non-negative, the second term is called the evidence lower bound on the marginal likelihood of data $(\mathbf{x}, \mathbf{y})$. This can be written as
\begin{equation}
    \begin{aligned}
        \log p(\mathbf{x}, \mathbf{y}) &\geq \mathbb{E}_q \log \frac{p(\mathbf{x}, \mathbf{y}, \mathbf{z}^c, \mathbf{z}^w, \mathbf{z}^v)}{q (\mathbf{z}^c, \mathbf{z}^w, \mathbf{z}^v | \mathbf{x}, \mathbf{y})} \\
        &= \mathbb{E}_{\mathbf{z}^c,\mathbf{z}^w,\mathbf{z}^v} \big[ \log p(\mathbf{x}|\mathbf{z}^c,\mathbf{z}^w)p(\mathbf{y}|\mathbf{z}^c,\mathbf{z}^v) \big] \\
        &- \mathbb{D}_{KL}(q(\mathbf{z}^c|\mathbf{x}),p(\mathbf{z}^c)) - \mathbb{D}_{KL}(q(\mathbf{z}^w|\mathbf{x}),p(\mathbf{z}^w)) \\
        &- \mathbb{D}_{KL}(q(\mathbf{z}^v|\mathbf{y}),p(\mathbf{z}^v)).
    \end{aligned}
\end{equation}
This completes the proof. 

Theorem~\ref{theorem:static} shows that for a specified domain, the latent space can be decoupled into a domain invariant subspace, a domain related subspace and a category related subspace.


\noindent\textbf{Probabilistic model for non-stationary environments.}~We now extend the aforementioned analysis to the non-stationary environment. Specifically, for the non-stationary environment, we can cast the dynamic variational inference framework via modeling the sequence of latent variables $\mathbf{z}^w$ and $\mathbf{z}^v$ as two parallel Markov chains~(i.e., $p(\mathbf{z}^w)=p(\mathbf{z}^w_t|\mathbf{z}^w_{<t})$ and $p(\mathbf{z}^v)=p(\mathbf{z}^v_t|\mathbf{z}^v_{<t})$). We thus have the following theorem.

\begin{theorem}
By denoting $p(\mathbf{z}^w)=p(\mathbf{z}^w_t|\mathbf{z}^w_{<t})$ and $p(\mathbf{z}^v)=p(\mathbf{z}^v_t|\mathbf{z}^v_{<t})$), $\mathcal{L}_{LSSAE}$ is equivalent to the ELBO of the data log likelihood $\log p(\mathbf{x}_{1:T}, \mathbf{y}_{1:T})$ based on the  non-stationary environment setting.
\end{theorem}

\emph{Proof.}~We can reformulate the lower bound in Eq.~\ref{equ:static_elbo} for a non-stationary environment as
\begin{equation}
\label{equ:dynamic_log}
    \log p(\mathbf{x}_{1:T}, \mathbf{y}_{1:T}) \geq 
    \mathbb{E}_q \log \frac{p(\mathbf{x}_{1:T}, \mathbf{y}_{1:T}, \mathbf{z}^c, \mathbf{z}^w_{1:T}, \mathbf{z}^v_{1:T})}{q (\mathbf{z}^c, \mathbf{z}^w_{1:T}, \mathbf{z}^v_{1:T} | \mathbf{x}_{1:T}, \mathbf{y}_{1:T})},
\end{equation}
which can  be written as
\begin{equation}
\label{equ:dynamic_elbo}
    \begin{aligned}
        \log~&p(\mathbf{x}_{1:T},\mathbf{y}_{1:T}) \\
        & \geq \sum_{t=1}^T \mathbb{E}_{\mathbf{z}^c,\mathbf{z}^w_t,\mathbf{z}^v_t} \big[ \log p(\mathbf{x}_t|\mathbf{z}^c,\mathbf{z}^w_t)p(\mathbf{y}|\mathbf{x}^c,\mathbf{z}^v_t) \big] \\
        &- \mathbb{D}_{KL}(q(\mathbf{z}^c|\mathbf{x}_{1:T}), p(\mathbf{z}^c)) \\
        &- \displaystyle\sum_{t=1}^T \mathbb{D}_{KL}(q(\mathbf{z}^w_t|\mathbf{z}^w_{<t},\mathbf{x}_t), p(\mathbf{z}^w_t|\mathbf{z}^w_{<t})) \\
        &- \displaystyle\sum_{t=1}^T \mathbb{D}_{KL}(q(\mathbf{z}^v_t|\mathbf{z}^v_{<t}, \mathbf{y}_t), p(\mathbf{z}^v_t|\mathbf{z}^v_{<t})),
    \end{aligned}
\end{equation}
where $\mathbf{z}^c \sim q(\mathbf{z}^c|\mathbf{x}_{1:T}), \mathbf{z}^w_t \sim q(\mathbf{z}^w_t|\mathbf{z}^w_{<t},\mathbf{x}_t)$ and $\mathbf{z}^v_t \sim q(\mathbf{z}^v_t|\mathbf{z}^v_{<t}, \mathbf{y}_t)$. The detailed derivation procedure is provided in App.~\ref{app:proofs}. We can see that the reconstruction part for $\mathbf{x}_t$ in the first term together with the second and third terms can form our objective $\mathcal{L}_d$ for domain-related module (i.e., covariate shift), and the combination of the reconstruction part for $\mathbf{y}_t$ in the first term and the last term can form $\mathcal{L}_c$ for category-related module (i.e., concept shift). As a result, the formulation above is equivalent to our objective $\mathcal{L}_{LSSAE}$.


\begin{figure}[!tbp]
\begin{center}
\includegraphics[width=0.5\textwidth]{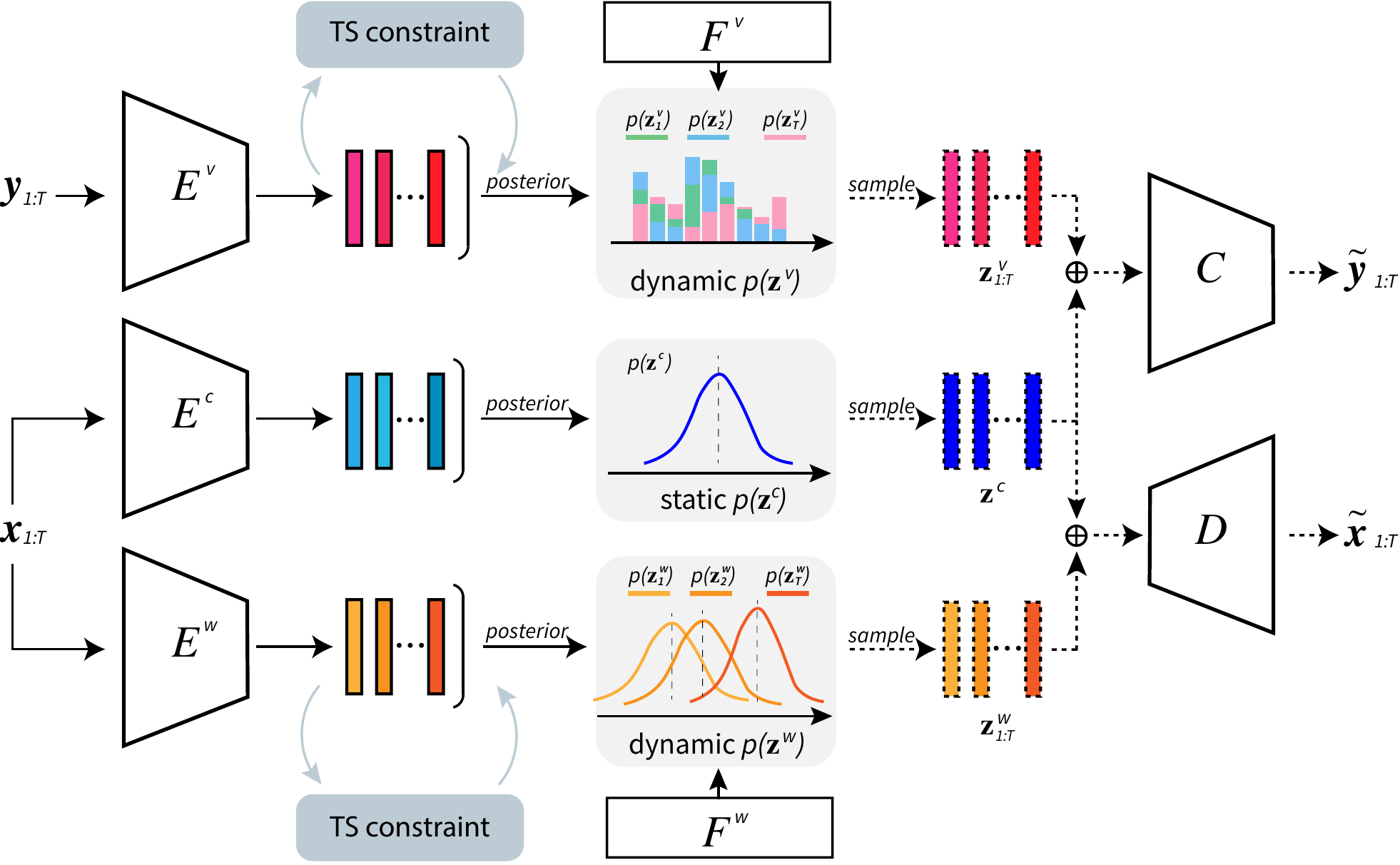}
\end{center}
\vspace{-0.3cm}
\caption{An overview of network architecture for LSSAE.
{Our framework consists of the static variational encoding network $E^c$, dynamic variational encoding networks $E^w$ and $E^v$, dynamic prior networks $F^w$ and $F^v$, a decoder $D$ and a classifier $C$.} It is worth noting that we do not require $E^v$ (i.e., only data from target domain $\mathcal{T}$ available) during inference stage.}
\label{fig:framework}
\end{figure}

\subsection{Implementation}
\label{sec:imp}

The implementation of network architecture for LSSAE is depicted in Fig.~\ref{fig:framework}. It is composed of two parts: (1) the domain-related module~(middle and bottom region); (2) the category-related module~(top region). 

The domain-related module consists of a static variational encoding network $E^c$ for $q(\mathbf{z}^c|\mathbf{x}_{1:T})$, a dynamic variational encoding network $E^w$ associated with $q(\mathbf{z}_t^w|\mathbf{z}_{<t}^w, \mathbf{x}_t)$, a dynamic prior network $F^w$ working for $p(\mathbf{z}_{t}^w|\mathbf{z}_{<t}^w)$ and a decoder $D$ corresponding to $p(\mathbf{x}_{t}|\mathbf{z}^c,\mathbf{z}_{t}^w)$. Similar to VAE \cite{kingma2014VAE},  we can apply the reparameterization trick~\cite{kingma2014VAE} to optimize parameters of $E^c$, $E^w$ and $F^w$. Specifically, we implement $E^c$ by a feature extractor which will be also utilized to extract features during test time. $E^w$ can be implemented by a feature extractor with same architecture of $E^c$ but not sharing network parameters, and followed by a LSTM network. $F^w$ is implemented as a one-layer LSTM network.  For the category-related module, we design a dynamic inference network $E^w$ which takes the one-hot code of label $\mathbf{y}_t$ as the input and with the output of the categorical distribution $q(\mathbf{z}^v_t|\mathbf{z}_{<t}^v,\mathbf{y}_t)$, and a classifier $C$ which takes $\mathbf{z}^c$ and $\mathbf{z}^v_t$ as the input for $p(\mathbf{y}_{t}|\mathbf{z}^c,\mathbf{z}_{t}^v)$. Similarly, the prior network $F^v$ for $p(\mathbf{z}^v_t|\mathbf{z}^v_{<t})$ is a LSTM network with a categorical distribution as the output. After that, we use Gumbel-Softmax reparameterization trick~\cite{Jang2017Gumbel,Maddison2017Gumbel} to sample $\mathbf{z}_t^v$ for optimization. Regarding the classifier $C$, we utilize a linear layer by following~\citet{Gulrajani2021ICLR}. The details of our architecture and hyperparameters for objective function can be found in the supplementary material. 

\noindent{\textbf{Optimization.}}~
{We first initialize $\mathbf{z}^w$ and $\mathbf{z}^v$ as $\mathbf{z}^w_0=\mathbf{0}$ and  $\mathbf{z}^v_0=\mathbf{0}$, respectively. To train our framework, we generate the dynamic prior distributions $p(\mathbf{z}^w_t|\mathbf{z}^w_{<t})$ and $p(\mathbf{z}^v_t|\mathbf{z}^v_{<t})$ through $F^w$ and $F^v$ respectively based on $T$ source domains $\mathcal{S}$.} For time stamp $t$, we sample a batch of data $\mathbf{x}_t$ and take it as input for $E^c$ and $E^w$ to obtain the parameters of their posterior distributions. After that, the latent features $\mathbf{z}^c$ and $\mathbf{z}^w_t$ are resampled separately through reparameterization trick and then concatenated together. Finally the decoder $D$ outputs the reconstruction data of $\mathbf{x}_t$. Meanwhile, $E^v$ takes the corresponding labels $\mathbf{y}_t$ of $\mathbf{x}_t$ in one-hot format as input and outputs the latent features, and then the latent features are also resampled through reparameterization trick to obtain $\mathbf{z}^v_t$. The classifier $C$ takes $\mathbf{z}^c$ and $\mathbf{z}^v_t$ as the input to predict the labels of $\mathbf{x}_t$. During training, we sample a mini-batch from each single domain with the same number of data sample to form a large batch in order to suit our framework to the temporal smooth constraint for stable training. This optimization procedure of LSSAE is depicted in Algorithm~\ref{alg:optimization}.

\noindent{\textbf{Inference.}}~ To predict the label of $\mathbf{x}_t$ sampled from one of the target domains in $\mathcal{D}_t$ in $\mathcal{T}$, we adopt $F^v$ to infer the latent code $\mathbf{z}^v_t$ and $E^c$ to extract the latent features $\mathbf{z}^c$, and then apply the classifier $C$ for the prediction purpose. It is worth noting that we do not require $E^v$ (i.e., only data from target domain $\mathcal{T}$ available) during inference stage. We show this procedure in Algorithm~\ref{alg:inference}.

\section{Experiments}
In this section, we present experimental results to validate the effectiveness of our proposed LSSAE based on the setting of evolving domain generalization. 

\subsection{Experimental Setup}
We compare the proposed LSSAE with other DG methods on two synthetic datesets~(Circle and Sine) and four real-world datasets~(Rotated MNIST, Portraits, Caltran, PowerSupply). We also evaluate the results on two variants named Circle-C and Sine-C derived from Circle and Sine via synthesizing concept shift manually. We split the domains into source domains, intermediate domains and target domains with the ratio of $\{1/2:1/6:1/3\}$. The intermediate domains are utilized as validation set. More details of dataset construction can be found in the supplementary material.



(1) \textbf{Circle/-C}~\cite{Pesaranghader2016FastHD}.~This dataset contains evolving 30 domains where the instance are sampled from 30 2D Gaussian distributions.  For {Circle-C}, concept shift is introduced via changing the center and radius of decision boundary in a gradual manner over time. 

(2) \textbf{Sine/-C}~\cite{Pesaranghader2016FastHD}.~We rearrange this dataset by extending it to  24 evolving domains. To simulate concept drift for {Sine-C}, labels are reversed (i.e., from 0 to 1 or from 1 to 0) from the $6$-th domain to the last one.

\begin{figure}[!tbp]
\vspace*{-\baselineskip}
\begin{minipage}{\columnwidth}
\begin{algorithm}[H]
   \caption{Optimization procedure for LSSAE}
   \label{alg:optimization}
\begin{algorithmic}
   \STATE {\bfseries Input:} sequential source labeled datasets $\mathcal{S}$; static feature extractor $E^c$; dynamic inference networks $E^{w}$, $E^{v}$ and their corresponding prior networks $F^{w}$, $F^{v}$; decoder $D$ and classifier $C$.
   \STATE Randomly initialize $E^c,E^{w},E^{v},F^{w},F^{v},D,C$
   \STATE Assign $\mathbf{z}^w_0, \mathbf{z}^v_0 \gets \mathbf{0}$

   \FOR {$t = 1,2,...,K$}
   \STATE Generate prior distribution $p(\mathbf{z}_{t}^w|\mathbf{z}_{<t}^w)$ via $F^{w}$
  \STATE Generate prior distribution $p(\mathbf{z}_{t}^v|\mathbf{z}_{<t}^v)$ via $F^{v}$
      \FOR {$i = 1, 2, ...$}
      \STATE Sample a batch of data $(\mathbf{x}_t, \mathbf{y}_t)$ from $\mathcal{S}_t$
      \STATE $\triangleright$ Calculate $\mathcal{L}_d$ by Eq.~\ref{equ:loss_d} for $E^c$, $E^{w}$, $D$ and $F^{w}$
      
      \STATE $\triangleright$ Calculate $\mathcal{L}_c$ by Eq.~\ref{equ:loss_c} for $E^c$, $E^{v}$, $C$ and $F^{v}$

      \STATE $\triangleright$ Calculate temporal smooth constriction by Eq.~\ref{equ:ts}
      Update all modules by the summary of these loss 
      \ENDFOR
   \ENDFOR
\end{algorithmic}
\end{algorithm}
\end{minipage}

\begin{minipage}{\columnwidth}
\begin{algorithm}[H]
   \caption{Inference procedure for LSSAE}
   \label{alg:inference}
\begin{algorithmic}
   \STATE {\bfseries Input:} sequential target datasets $\mathcal{T}$; static feature extractor $E^c$; dynamic prior network $F^{v}$ and classifier $C$.
   \STATE Assign $\mathbf{z}^v_0 \gets \mathbf{0}$

   \FOR {$t = 1,2,...$}
   \STATE Sample $\mathbf{z}^v_t \sim p(\mathbf{z}_{t}^v|\mathbf{z}_{<t}^v)$ via $F^{v}$
      \FOR {$i = 1, 2, ...$}
      \STATE $\triangleright$ Extract the feature for data $\mathbf{x}_t$ via $E^c$
      \STATE $\triangleright$ Generate the prediction $\mathbf{\tilde{y}}_t$ via $C$
      \ENDFOR
   \ENDFOR
\end{algorithmic}
\end{algorithm}
\end{minipage}
\end{figure}

(3) \textbf{Rotated MNIST (RMNIST)}~\cite{Ghifary2015RMNIST}.~Rotated MNIST (RMNIST) is composed of MNIST digits of various rotations. We extend it to 19 evolving domains via applying the rotations with degree of $\mathcal{R}=\{0^\circ,15^\circ,30^\circ,...,180^\circ\}$ in order. 

(4) \textbf{Portraits}~\cite{Ginosar2015ICCVW}.~This dataset comprises photos of high-school seniors from the 1905s to the 2005s for gender classification. We split the dataset into 34 domains by a fixed internal over time. 

(5) \textbf{Caltran}~\cite{Hoffman2014CVPR}.~Caltran is a real-world surveillance dataset comprising images captured from a fixed traffic camera deployed in an intersection. The task is to predict the type of scene based on continuously evolving data. We divide it into 34 domains based on different times.

\begin{table*}[!tbp]
\caption{The comparison of accuracy~(\%) between LSSAE and other DG baselines on various datasets. As DG baselines do not take concept shift into consideration, we report the results on datasets without concept shift and with concept shift separately.}
\label{tab:results}
\vskip 0.15in
\begin{center}
\begin{small}
\begin{tabular}{lcccccc|cccc}
\toprule
\textbf{Algorithm} & \textbf{Circle} & \textbf{Sine} & \textbf{RMNIST} &  \textbf{Portraits} & \textbf{Caltran} & \textcolor{teal}{\textbf{Avg}} & \textbf{Circle-C} & \textbf{Sine-C} & \textbf{PowerSupply} & \textcolor{teal}{\textbf{Avg}}\\
\midrule
ERM     & 49.9 & 63.0 & 43.6 & 87.8 & 66.3 & 62.1 & 34.0 & 61.5 & 71.0 & 55.5 \\
Mixup   & 48.4 & 62.9 & 44.9 & 87.8 & 66.0 & 62.0 & 33.9 & 60.9 & 70.8 & 55.2 \\
MMD     & 50.7 & 55.8 & 44.8 & 87.3 & 57.1 & 59.1 & 33.7 & 52.7 & 70.9 & 52.4 \\
MLDG    & 50.8 & 63.2 & 43.1 & 88.5 & 66.2 & 62.3 & 34.6 & 62.0 & 70.8 & 55.8 \\
IRM     & 51.3 & 63.2 & 39.0 & 85.4 & 64.1 & 60.6 & 38.5 & 61.2 & 70.8 & 56.8 \\
RSC     & 48.0 & 61.5 & 41.7 & 87.3 & 67.0 & 61.1 & 33.7 & 61.5 & 70.9 & 55.4 \\
MTL     & 51.2 & 62.9 & 41.7 & 89.0 & 68.2 & 62.6 & 33.9 & 61.4 & 70.7 & 55.3 \\
Fish    & 48.8 & 62.3 & 44.2 & 88.8 & 68.6 & 62.5 & 34.3 & \textbf{62.7} & 70.8 & 55.9\\
CORAL   & 53.9 & 51.6 & 44.5 & 87.4 & 65.7 & 60.6 & 34.1 & 59.0 & 71.0 & 54.7 \\
AndMask & 47.9 & 69.3 & 42.8 & 70.3 & 56.9 & 57.4 & 37.7 & 52.7 & 70.7 & 53.7 \\
DIVA    & 67.9 & 52.9 & 42.7 & 88.2 & 69.2 & 64.2 & 33.9 & 52.9 & 70.8 & 55.1 \\
LSSAE~(Ours) & \textbf{73.8} & \textbf{71.4} & \textbf{46.4} & \textbf{89.1} & \textbf{70.6} & \textbf{70.3} & \textbf{44.8} & 60.8 & \textbf{71.1} & \textbf{58.9} \\
\bottomrule
\end{tabular}
\end{small}
\end{center}
\vskip -0.1in
\end{table*}

(6) \textbf{PowerSupply}~\cite{Dau2019UCR}.~ PowerSupply is constructed for the time-section prediction of current power supply based on the hourly records of an Italy electricity
company. The concept shift may raise from the change in season, weather or price. We form $30$ domains according to days.

The methods for comparison include: (1)~ERM~\cite{Vapnik1998ERM}; (2)~Mixup~\cite{Yan2020Mixup}; (3)~MMD~\cite{Li2018MMD}; (4)~MLDG~\cite{Li2018AAAI}; (5)~IRM~\cite{Rosenfeld2021IRM}; (6)~RSC~\cite{Huang2020RSC}; (7)~MTL~\cite{Blanchard2021MTL}; (8)~Fish~\cite{Shi2021Fish};
(9)~CORAL~\cite{Sun2016CORAL}; (10)~AndMask~\cite{Parascandolo2021ANDMask}; (11)~DIVA~\cite{Ilse2020Diva}. 
All of our experiments are implemented in the PyTorch platform based on DomainBed package~\cite{Gulrajani2021ICLR}. For a fair comparison, we keep the neural network architecture of encoding part and classification part to be same for all baselines for different benchmarks.

\subsection{Quantitative Results}
The results of of our proposed LSSAE and baselines are presented in Table~\ref{tab:results}. As conventional DG methods focus upon covariate shift only, we separate the datasets according to with or without concept shift into two parts for fairness. We can see that LSSAE consistently outperforms other baselines over all datasets, it achieves $70.3\%$ accuracy when there exists covariate shift only~(Circle, Sine, RMNIST, Portraits and Caltran), and achieves $58.9\%$ accuracy when there exist concept shift~(Circle-C, Sine-C, PowerSupply). The results are significantly better than the compared DG approaches, which are reasonable since existing DG methods cannot deal with distribution shift well in non-stationary environments but our proposed LSSAE can properly capture the evolving patterns to gain better performance. More results can be found in the supplementary materials. 

\begin{figure}[!tbp]
	\centering
	\begin{minipage}{\linewidth}
		\centering
		\includegraphics[width=0.95\linewidth]{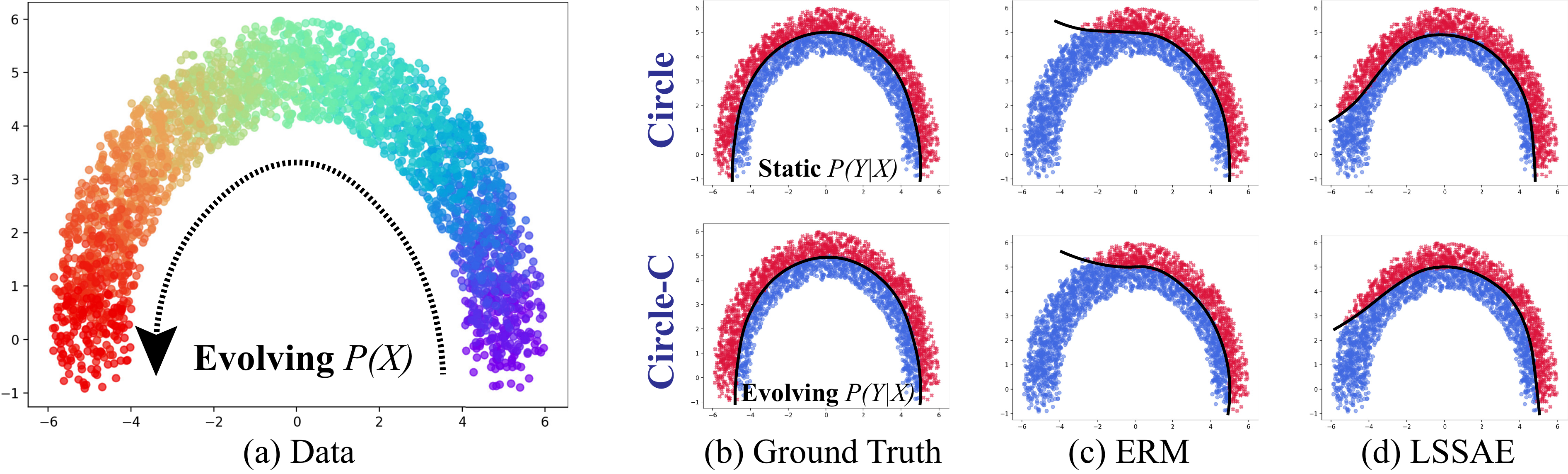}
        \vspace{-0.3cm}
		\caption{Decision boundary visualization of Circle and Circle-C datasets each with 30 domains. (a) presents the original data in different domains by color, where the right half part are source domains. (b) shows the positive and negative labels in red and blue dots. (c) and (d) are decision boundaries learned by ERM and LSSAE, respectively.}
		\label{fig:circle}
	\hfill
	\end{minipage}
    
	\begin{minipage}{\linewidth}
		\centering
		\includegraphics[width=0.95\linewidth]{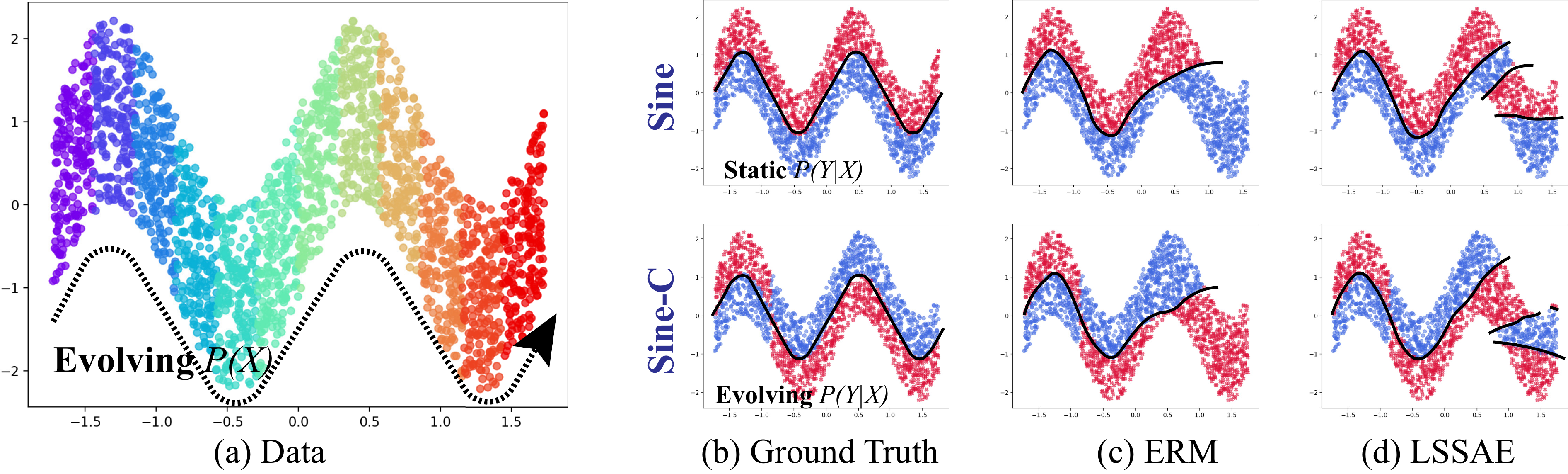}
        \vspace{-0.3cm}
		\caption{Decision boundary visualization of Sine and Sine-C datasets. (a) presents 24 domains indexed by different colors, where the left half part are source domains. (b) shows the positive and negative labels in red and blue dots. (c) and (d) are decision boundaries learned by ERM and LSSAE, respectively.}
		\label{fig:sine}
	\end{minipage}
\end{figure}

To better understand the rationality of our method, we visualize the decision boundaries of our method and ERM baseline on two synthetic datasets: Circle and Sine by comparing our proposed method with ERM baseline.  The visualization results are depicted in Fig.~\ref{fig:circle} and Fig.~\ref{fig:sine}. As shown in Fig.~\ref{fig:circle}, both of ERM and LSSAE can fit the source domains well. However, different from ERM which only  fit the source domains, LSSAE shows a desired generalization ability to unseen target domains. This validates that our LSSAE can capture the underlying evolving patterns across domains to achieve better results. As for Circle-C where we introduce evolving concept shift via modifying the center and radius of decision boundary over a period of time, we observe that our proposed LSSAE can still produce a more accurate decision boundary compared with ERM. Similar observation can be found in Fig.~\ref{fig:sine}, where LSSAE can be better generalized to evolving domains compared with ERM. 
However, we find that our proposed LSSAE may not be able to obtain a desired boundary based on the setting of Sine-C, where we introduce concept shift by reversing the labels. We conjecture the reason that our proposed LSSAE may not be able to well handle the abrupt concept shift where no continuous evolving pattern exists.


\subsection{Ablation Study}
\label{sec:ablation_study}

\noindent\textbf{Analysis on domain-related module.}
\begin{figure}[!tbp]
    \begin{center}
    \includegraphics[width=0.46\textwidth]{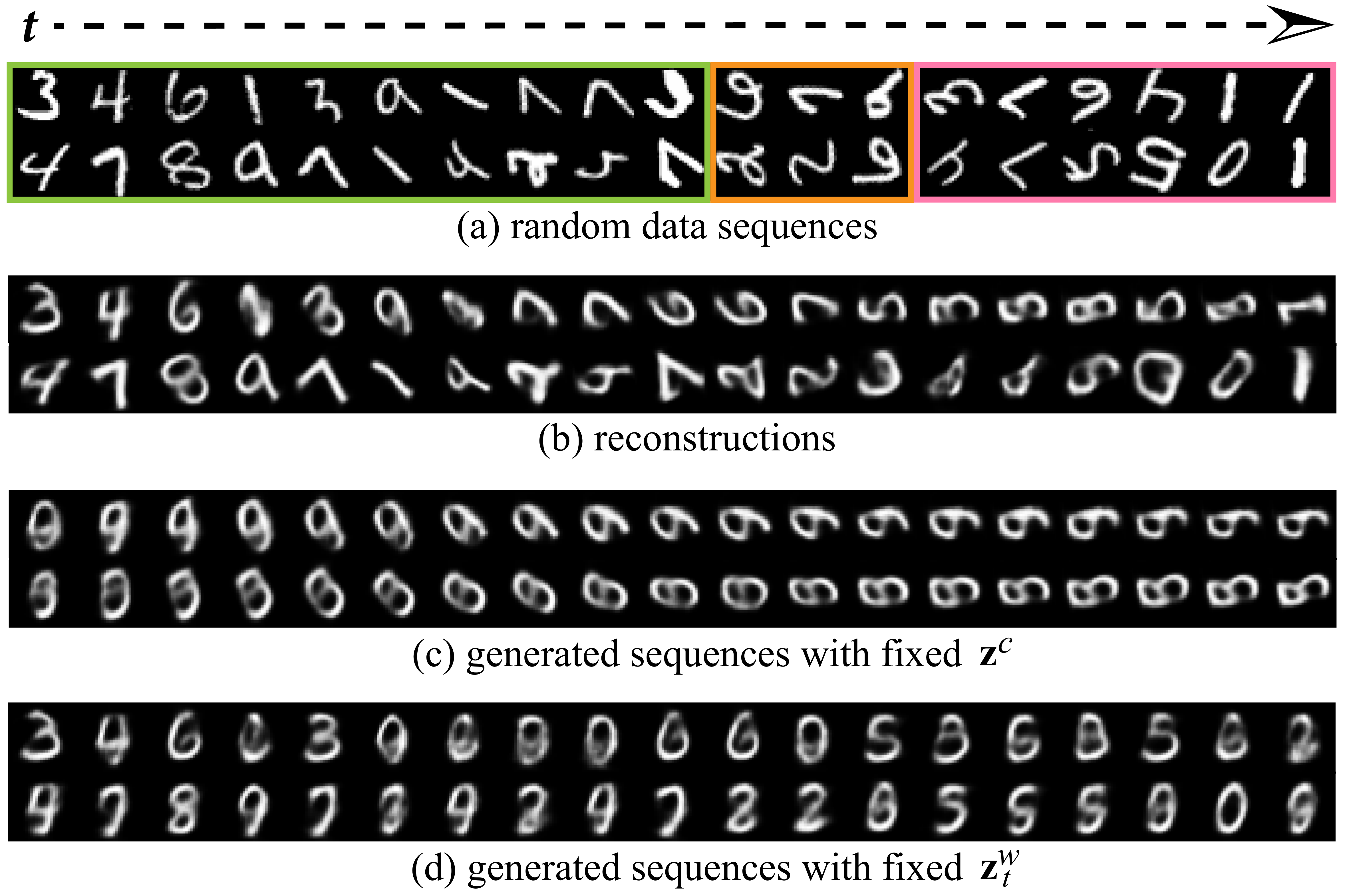}
    \end{center}
    \vspace{-0.3cm}
    \caption{Visualisation of generated and reconstructed data sequences on RMNIST dataset.} 
    \label{fig:gen}
\end{figure}
To better evaluate whether our proposed LSSAE can capture domain specific information, we conduct experiments by evaluating the reconstruction and generation capability of LSSAE on RMNIST dataset. The reconstruction and generation results are shown in Fig.~\ref{fig:gen}. Each subfigure shows two sequences which follow the domain order (i.e., rotation degree gradually evolves) from the left to the right. Subfigure~(a) shows the original data sequence sampled from source domains~(green bounding box), intermediate domains~(orange bounding box) and target domains~(pink bounding box) in order. Noted that we are only interested in degree of rotation thus the category of images from different domains may not be consistent. Subfigure~(b) shows the reconstructions for each sample in Subfigure~(a). Based on the results, we find that we can achieve a desired reconstruction quality, which suggests that the domain-related information can be captured even we random the category related information. To further show that our proposed method is capable for generation task, in subfigure~(c), we visualize the randomly generated samples using $\mathbf{z}_{t}^w \sim p(\mathbf{z}_{t}^w|\mathbf{z}_{<t}^w)$ while keeping  $\mathbf{z}^c$ the same for all domains. We can see that the category information (i.e., digit information) remains unchanged when we fix $\mathbf{z}^c$, and can generated samples for unseen future domains where the rotation degree of digit is gradually changed with evolving prior $p(\mathbf{z}_{t}^w|\mathbf{z}_{<t}^w)$. This observation suggests that our proposed method can be used for data augmentation when adapting to the unseen target domains. However, we also observe that there exists image quality distortion when generating new domains. We conjecture the reason that the range of degrees for training may not be sufficiently diverse, which limits the performance of generation to unseen digit rotation degrees. Subfigure~(d) shows randomly generated data via sampling $\mathbf{z}^c \sim p(\mathbf{z}^c)$ while keeping $\mathbf{z}^w_t$ the same for all domains at different time stamp $t$~(\ie~$\mathbf{z}^w_t=\mathbf{z}^w_1$), where we find that the generated digit images belong to different categories but with the same rotation degree, which further indicates that our proposed method can successfully extract domain related information.


\noindent\textbf{Analysis on category-related module.}
\begin{table}[!tbp]
\caption{Comparison of different prior distributions for category-related module on PowerSupply dataset.}
\label{tab:abalation_concept}
\vskip 0.15in
\begin{center}
\begin{small}
\begin{tabular}{p{100pt}<{\raggedright} p{70pt}<{\centering}}
\toprule
\textbf{Prior Type} & \textbf{Accuracy (\%)} \\
\midrule
Without $\mathbf{z}^v$  & 70.7  \\
Gaussian                & 70.1  \\
Uniform                 & 71.0  \\
Categorical~(Ours)      & \textbf{71.1}  \\
\bottomrule
\end{tabular}
\end{small}
\end{center}
\vskip -0.1in
\end{table}
We then evaluate the effectiveness of our proposed category-related module on PowerSupply dataset. To this end, we consider three different ablation studies by 1) removing category-related module, 2) replacing the learnable categorical prior with a learnable Gaussian prior, 3) replacing the learnable categorical prior with a fix Uniform prior. The results are shown in Table~\ref{tab:abalation_concept}. As we can see, our proposed LSSAE based on categorical prior can achieve the best performance among other baselines, which shows the effectiveness of our proposed method. However, we observe that the performance drops to some extent by replacing the categorical prior with Gaussian prior, which we conjecture the reason that categorical distribution can better capture the category related information which is discrete. We also observe that better performance can be achieved by comparing with the results using Uniform distribution as prior, which is reasonable since Uniform distribution does not change over time. 

\noindent\textbf{Analysis on temporal smooth constraint.}~
The smooth constraint is mainly designed for stabilizing the training procedure. To verify this, we conduct experiments by considering ``with \& without this constraint" on RMNIST and CalTran datasets while keeping other hyperparameters the same. The results are reported as the variance of the test accuracy in the last five epochs (See Table~\ref{tab:ablation_ts}). 
For RMNIST, when training with constraint, the variance is $0.4$; without constraint, the variance is $3.6$. For CalTran, when training with constraint, the variance is $4.7$, and without constraint, the variance is $10.0$. We can see that there exists an obvious decrease in terms of variance when incorporating with our proposed TS constraint. Besides, as our smooth constraint helps stabilize the training process, better performance can be expected. For RMNIST, training with our smooth constraint can achieve $0.9\%$ performance gain; For CalTran, the performance gain is $0.7\%$.
\begin{table}[!tbp]
\caption{Ablation study on temporal smooth constraint.}
\label{tab:ablation_ts}
\vskip 0.15in
\begin{center}
\begin{small}
\begin{tabular}{ccccc}\hline
    \toprule
    \multirow{2}*{\textbf{TS Constraint}} &  \multicolumn{2}{c}{\textbf{RMNIST}} &  \multicolumn{2}{c}{\textbf{CalTran}} \\
    ~ &  \textbf{Var} $\downarrow$ &  \textbf{Acc (\%)} $\uparrow$ &  \textbf{Var} $\downarrow$ &  \textbf{Acc (\%)} $\uparrow$ \\
    \midrule
    \normalsize$\times$     & 3.6  & 45.5  & 10.0 & 69.9 \\
    \normalsize\checkmark   & 0.4 & 46.4   & 4.7 & 70.6 \\
    \bottomrule[1pt]
\end{tabular}
\end{small}
\end{center}
\vskip -0.1in
\end{table}

\section{Conclusion}
In this paper, we propose to focus on the problem of evolving domain generalization, where the covariate shift and concept shift vary over time. To tackle this problem, we propose a novel framework LSSAE to model the dynamics of distribution shift (i.e., covariate shift and concept shift). We also provide theoretical analysis, which shows that our proposed method is equivalent to maximizing the ELBO based on the non-stationary environment setting, and justifies the rationality of our proposed method for the problem of evolving domain generalization. Experimental results on both toy data and real-world datasets across multiple domains 
further indicate the significance of our proposed method based on this setting.

\section*{Acknowledgements}
This work was supported in part by CityU New Research Initiatives/Infrastructure Support from Central (APRC 9610528), CityU Applied Research Grant (ARG 9667244) and Hong Kong Innovation and Technology Commission (InnoHK Project CIMDA). Besides, this work was also supported in part by the National Natural Science Foundation of China under 62022002, in part by the Hong Kong Research Grants Council General Research Fund (GRF) under Grant 11203220.



\bibliography{egbib}

\begin{thebibliography}{49}
\providecommand{\natexlab}[1]{#1}
\providecommand{\url}[1]{\texttt{#1}}
\expandafter\ifx\csname urlstyle\endcsname\relax
  \providecommand{\doi}[1]{doi: #1}\else
  \providecommand{\doi}{doi: \begingroup \urlstyle{rm}\Url}\fi

\bibitem[Albuquerque et~al.(2021)Albuquerque, Monteiro, Darvishi, Falk, and
  Mitliagkas]{Albuquerque2021arXiv}
Albuquerque, I., Monteiro, J., Darvishi, M., Falk, T.~H., and Mitliagkas, I.
\newblock Generalizing to unseen domains via distribution matching.
\newblock \emph{arXiv preprint arXiv:1911.00804}, 2021.

\bibitem[Balaji et~al.(2018)Balaji, Sankaranarayanan, and
  Chellappa]{Balaji2018NIPS}
Balaji, Y., Sankaranarayanan, S., and Chellappa, R.
\newblock Metareg: Towards domain generalization using meta-regularization.
\newblock \emph{Advances in Neural Information Processing Systems},
  31:\penalty0 998--1008, 2018.

\bibitem[Blanchard et~al.(2011)Blanchard, Lee, and Scott]{Blanchard2011NIPS}
Blanchard, G., Lee, G., and Scott, C.
\newblock Generalizing from several related classification tasks to a new
  unlabeled sample.
\newblock \emph{Advances in Neural Information Processing Systems},
  24:\penalty0 2178--2186, 2011.

\bibitem[Blanchard et~al.(2021)Blanchard, Deshmukh, Dogan, Lee, and
  Scott]{Blanchard2021MTL}
Blanchard, G., Deshmukh, A.~A., Dogan, {\"U}., Lee, G., and Scott, C.
\newblock Domain generalization by marginal transfer learning.
\newblock \emph{J. Mach. Learn. Res.}, 22:\penalty0 2--1, 2021.

\bibitem[Chen \& Chao(2021)Chen and Chao]{Chen2021NIPS}
Chen, H.-Y. and Chao, W.-L.
\newblock Gradual domain adaptation without indexed intermediate domains.
\newblock In \emph{Thirty-Fifth Conference on Neural Information Processing
  Systems}, 2021.

\bibitem[Dau et~al.(2019)Dau, Bagnall, Kamgar, Yeh, Zhu, Gharghabi,
  Ratanamahatana, and Keogh]{Dau2019UCR}
Dau, H.~A., Bagnall, A., Kamgar, K., Yeh, C.-C.~M., Zhu, Y., Gharghabi, S.,
  Ratanamahatana, C.~A., and Keogh, E.
\newblock The ucr time series archive.
\newblock \emph{IEEE/CAA Journal of Automatica Sinica}, 6\penalty0
  (6):\penalty0 1293--1305, 2019.

\bibitem[Dou et~al.(2019)Dou, Coelho~de Castro, Kamnitsas, and
  Glocker]{Dou2019NIPS}
Dou, Q., Coelho~de Castro, D., Kamnitsas, K., and Glocker, B.
\newblock Domain generalization via model-agnostic learning of semantic
  features.
\newblock \emph{Advances in Neural Information Processing Systems},
  32:\penalty0 6450--6461, 2019.

\bibitem[Federici et~al.(2021)Federici, Tomioka, and
  Forr{\'e}]{Federici2021Theory}
Federici, M., Tomioka, R., and Forr{\'e}, P.
\newblock An information-theoretic approach to distribution shifts.
\newblock In \emph{Thirty-Fifth Conference on Neural Information Processing
  Systems}, 2021.

\bibitem[Ghifary et~al.(2015)Ghifary, Kleijn, Zhang, and
  Balduzzi]{Ghifary2015RMNIST}
Ghifary, M., Kleijn, W.~B., Zhang, M., and Balduzzi, D.
\newblock Domain generalization for object recognition with multi-task
  autoencoders.
\newblock In \emph{Proceedings of the IEEE International Conference on Computer
  Vision}, pp.\  2551--2559, 2015.

\bibitem[Ginosar et~al.(2015)Ginosar, Rakelly, Sachs, Yin, and
  Efros]{Ginosar2015ICCVW}
Ginosar, S., Rakelly, K., Sachs, S., Yin, B., and Efros, A.~A.
\newblock A century of portraits: A visual historical record of american high
  school yearbooks.
\newblock In \emph{2015 IEEE International Conference on Computer Vision
  Workshop (ICCVW)}, pp.\  652--658, 2015.

\bibitem[Gong et~al.(2019)Gong, Li, Chen, and Gool]{Gong2019CVPR}
Gong, R., Li, W., Chen, Y., and Gool, L.~V.
\newblock Dlow: Domain flow for adaptation and generalization.
\newblock In \emph{Proceedings of the IEEE/CVF Conference on Computer Vision
  and Pattern Recognition}, pp.\  2477--2486, 2019.

\bibitem[Gulrajani \& Lopez-Paz(2021)Gulrajani and
  Lopez-Paz]{Gulrajani2021ICLR}
Gulrajani, I. and Lopez-Paz, D.
\newblock In search of lost domain generalization.
\newblock In \emph{International Conference on Learning Representations}, 2021.

\bibitem[Han et~al.(2021)Han, Min, Han, Li, and Zhang]{Han2021RWAE}
Han, J., Min, M.~R., Han, L., Li, L.~E., and Zhang, X.
\newblock Disentangled recurrent wasserstein autoencoder.
\newblock In \emph{International Conference on Learning Representations}, 2021.

\bibitem[Hochreiter \& Schmidhuber(1997)Hochreiter and
  Schmidhuber]{Hochreiter1997LSTM}
Hochreiter, S. and Schmidhuber, J.
\newblock Long short-term memory.
\newblock \emph{Neural Computation}, 9\penalty0 (8):\penalty0 1735--1780, 1997.

\bibitem[Hoffman et~al.(2014)Hoffman, Darrell, and Saenko]{Hoffman2014CVPR}
Hoffman, J., Darrell, T., and Saenko, K.
\newblock Continuous manifold based adaptation for evolving visual domains.
\newblock In \emph{Proceedings of the IEEE Conference on Computer Vision and
  Pattern Recognition}, pp.\  867--874, 2014.

\bibitem[Huang et~al.(2020)Huang, Wang, Xing, and Huang]{Huang2020RSC}
Huang, Z., Wang, H., Xing, E.~P., and Huang, D.
\newblock Self-challenging improves cross-domain generalization.
\newblock In \emph{Computer Vision--ECCV 2020: 16th European Conference,
  Glasgow, UK, August 23--28, 2020, Proceedings, Part II 16}, pp.\  124--140,
  2020.

\bibitem[Ilse et~al.(2020)Ilse, Tomczak, Louizos, and Welling]{Ilse2020Diva}
Ilse, M., Tomczak, J.~M., Louizos, C., and Welling, M.
\newblock Diva: Domain invariant variational autoencoders.
\newblock In \emph{Medical Imaging with Deep Learning}, pp.\  322--348. PMLR,
  2020.

\bibitem[Jang et~al.(2017)Jang, Gu, and Poole]{Jang2017Gumbel}
Jang, E., Gu, S., and Poole, B.
\newblock Categorical reparameterization with gumbel-softmax.
\newblock \emph{International Conference on Learning Representations}, 2017.

\bibitem[Kingma \& Ba(2015)Kingma and Ba]{Kingma2015Adam}
Kingma, D.~P. and Ba, J.
\newblock Adam: A method for stochastic optimization.
\newblock In \emph{International Conference on Learning Representations}, 2015.

\bibitem[Kingma \& Welling(2014)Kingma and Welling]{kingma2014VAE}
Kingma, D.~P. and Welling, M.
\newblock Auto-encoding variational bayes.
\newblock \emph{International Conference on Learning Representations}, 2014.

\bibitem[Kumar et~al.(2020)Kumar, Ma, and Liang]{Kumar2020ICML}
Kumar, A., Ma, T., and Liang, P.
\newblock Understanding self-training for gradual domain adaptation.
\newblock In \emph{International Conference on Machine Learning}, pp.\
  5468--5479. PMLR, 2020.

\bibitem[Lao et~al.(2020)Lao, Jiang, Havaei, and Bengio]{Lao2020Continuous}
Lao, Q., Jiang, X., Havaei, M., and Bengio, Y.
\newblock Continuous domain adaptation with variational domain-agnostic feature
  replay.
\newblock \emph{arXiv preprint arXiv:2003.04382}, 2020.

\bibitem[Li et~al.(2018{\natexlab{a}})Li, Yang, Song, and
  Hospedales]{Li2018AAAI}
Li, D., Yang, Y., Song, Y.-Z., and Hospedales, T.~M.
\newblock Learning to generalize: Meta-learning for domain generalization.
\newblock In \emph{Thirty-Second AAAI Conference on Artificial Intelligence},
  2018{\natexlab{a}}.

\bibitem[Li et~al.(2018{\natexlab{b}})Li, Pan, Wang, and Kot]{Li2018MMD}
Li, H., Pan, S.~J., Wang, S., and Kot, A.~C.
\newblock Domain generalization with adversarial feature learning.
\newblock In \emph{Proceedings of the IEEE Conference on Computer Vision and
  Pattern Recognition}, pp.\  5400--5409, 2018{\natexlab{b}}.

\bibitem[Li et~al.(2017)Li, Xu, Xu, Dai, and Van~Gool]{Li2017PAMI}
Li, W., Xu, Z., Xu, D., Dai, D., and Van~Gool, L.
\newblock Domain generalization and adaptation using low rank exemplar svms.
\newblock \emph{IEEE Transactions on Pattern Analysis and Machine
  Intelligence}, 40\penalty0 (5):\penalty0 1114--1127, 2017.

\bibitem[Liu et~al.(2020)Liu, Long, Wang, and Wang]{Long2020EDA}
Liu, H., Long, M., Wang, J., and Wang, Y.
\newblock Learning to adapt to evolving domains.
\newblock In \emph{Advances in Neural Information Processing Systems},
  volume~33, pp.\  22338--22348, 2020.

\bibitem[Maddison et~al.(2017)Maddison, Mnih, and Teh]{Maddison2017Gumbel}
Maddison, C.~J., Mnih, A., and Teh, Y.~W.
\newblock The concrete distribution: A continuous relaxation of discrete random
  variables.
\newblock \emph{International Conference on Learning Representations}, 2017.

\bibitem[Mancini et~al.(2019)Mancini, Bulo, Caputo, and Ricci]{Mancini2019CVPR}
Mancini, M., Bulo, S., Caputo, B., and Ricci, E.
\newblock Adagraph: Unifying predictive and continuous domain adaptation
  through graphs.
\newblock In \emph{2019 IEEE/CVF Conference on Computer Vision and Pattern
  Recognition (CVPR)}, pp.\  6561--6570, jun 2019.

\bibitem[Muandet et~al.(2013)Muandet, Balduzzi, and
  Sch{\"o}lkopf]{Muandet2013ICML}
Muandet, K., Balduzzi, D., and Sch{\"o}lkopf, B.
\newblock Domain generalization via invariant feature representation.
\newblock In \emph{International Conference on Machine Learning}, pp.\  10--18,
  2013.

\bibitem[Nguyen et~al.(2021)Nguyen, Tran, Gal, and Baydin]{Nguyen2021NIPS}
Nguyen, A.~T., Tran, T., Gal, Y., and Baydin, A.~G.
\newblock Domain invariant representation learning with domain density
  transformations.
\newblock \emph{arXiv preprint arXiv:2102.05082}, 2021.

\bibitem[Parascandolo et~al.(2021)Parascandolo, Neitz, ORVIETO, Gresele, and
  Sch{\"o}lkopf]{Parascandolo2021ANDMask}
Parascandolo, G., Neitz, A., ORVIETO, A., Gresele, L., and Sch{\"o}lkopf, B.
\newblock Learning explanations that are hard to vary.
\newblock In \emph{International Conference on Learning Representations}, 2021.

\bibitem[Park et~al.(2021)Park, Shu, and Kwon]{Park2021ICML}
Park, S.~W., Shu, D.~W., and Kwon, J.
\newblock Generative adversarial networks for markovian temporal dynamics:
  Stochastic continuous data generation.
\newblock In \emph{Proceedings of the 38th International Conference on Machine
  Learning}, volume 139, pp.\  8413--8421, 18--24 Jul 2021.

\bibitem[Pesaranghader \& Viktor(2016)Pesaranghader and
  Viktor]{Pesaranghader2016FastHD}
Pesaranghader, A. and Viktor, H.~L.
\newblock Fast hoeffding drift detection method for evolving data streams.
\newblock In \emph{ECML/PKDD}, 2016.

\bibitem[Rosenfeld et~al.(2021)Rosenfeld, Ravikumar, and
  Risteski]{Rosenfeld2021IRM}
Rosenfeld, E., Ravikumar, P.~K., and Risteski, A.
\newblock The risks of invariant risk minimization.
\newblock In \emph{International Conference on Learning Representations}, 2021.

\bibitem[Shankar et~al.(2018)Shankar, Piratla, Chakrabarti, Chaudhuri, Jyothi,
  and Sarawagi]{Shankar2018ICLR}
Shankar, S., Piratla, V., Chakrabarti, S., Chaudhuri, S., Jyothi, P., and
  Sarawagi, S.
\newblock Generalizing across domains via cross-gradient training.
\newblock In \emph{International Conference on Learning Representations}, 2018.

\bibitem[Shi et~al.(2021)Shi, Seely, Torr, Siddharth, Hannun, Usunier, and
  Synnaeve]{Shi2021Fish}
Shi, Y., Seely, J., Torr, P.~H., Siddharth, N., Hannun, A., Usunier, N., and
  Synnaeve, G.
\newblock Gradient matching for domain generalization.
\newblock \emph{arXiv preprint arXiv:2104.09937}, 2021.

\bibitem[Sugiyama et~al.(2013)Sugiyama, Yamada, and
  du~Plessis]{Sugiyama2013WIRCS}
Sugiyama, M., Yamada, M., and du~Plessis, M.~C.
\newblock Learning under nonstationarity: covariate shift and class-balance
  change.
\newblock \emph{Wiley Interdisciplinary Reviews: Computational Statistics},
  5\penalty0 (6):\penalty0 465--477, 2013.

\bibitem[Sun \& Saenko(2016)Sun and Saenko]{Sun2016CORAL}
Sun, B. and Saenko, K.
\newblock Deep coral: Correlation alignment for deep domain adaptation.
\newblock In Hua, G. and J{\'e}gou, H. (eds.), \emph{Computer Vision -- ECCV
  2016 Workshops}, pp.\  443--450, Cham, 2016.

\bibitem[Tahmasbi et~al.(2021)Tahmasbi, Jothimurugesan, Tirthapura, and
  Gibbons]{Tahmasbi2021DriftSurf}
Tahmasbi, A., Jothimurugesan, E., Tirthapura, S., and Gibbons, P.~B.
\newblock Driftsurf: Stable-state/reactive-state learning under concept drift.
\newblock In \emph{International Conference on Machine Learning}, pp.\
  10054--10064. PMLR, 2021.

\bibitem[Torralba \& Efros(2011)Torralba and Efros]{Torralba2011CVPR}
Torralba, A. and Efros, A.~A.
\newblock Unbiased look at dataset bias.
\newblock In \emph{CVPR 2011}, pp.\  1521--1528. IEEE, 2011.

\bibitem[Vapnik(1998)]{Vapnik1998ERM}
Vapnik, V.
\newblock Statistical learning theory wiley.
\newblock \emph{New York}, 1998.

\bibitem[Vergara et~al.(2012)Vergara, Vembu, Ayhan, Ryan, Homer, and
  Huerta]{Vergara2012Chemical}
Vergara, A., Vembu, S., Ayhan, T., Ryan, M.~A., Homer, M.~L., and Huerta, R.
\newblock Chemical gas sensor drift compensation using classifier ensembles.
\newblock \emph{Sensors and Actuators B: Chemical}, 166:\penalty0 320--329,
  2012.

\bibitem[Volpi et~al.(2018)Volpi, Namkoong, Sener, Duchi, Murino, and
  Savarese]{Volpi2018NIPS}
Volpi, R., Namkoong, H., Sener, O., Duchi, J.~C., Murino, V., and Savarese, S.
\newblock Generalizing to unseen domains via adversarial data augmentation.
\newblock In \emph{Advances in Neural Information Processing Systems},
  volume~31, 2018.

\bibitem[Wang et~al.(2020)Wang, He, and Katabi]{Wang2020CIDA}
Wang, H., He, H., and Katabi, D.
\newblock Continuously indexed domain adaptation.
\newblock In III, H.~D. and Singh, A. (eds.), \emph{Proceedings of the 37th
  International Conference on Machine Learning}, volume 119, pp.\  9898--9907,
  13--18 Jul 2020.

\bibitem[Wang et~al.(2021)Wang, Li, Chau, and Kot]{Wang2021arXiv}
Wang, Y., Li, H., Chau, L.-P., and Kot, A.~C.
\newblock Variational disentanglement for domain generalization.
\newblock \emph{arXiv preprint arXiv:2109.05826}, 2021.

\bibitem[Wulfmeier et~al.(2018)Wulfmeier, Bewley, and
  Posner]{Wulfmeier2018ICRA}
Wulfmeier, M., Bewley, A., and Posner, I.
\newblock Incremental adversarial domain adaptation for continually changing
  environments.
\newblock In \emph{2018 IEEE International Conference on Robotics and
  Automation (ICRA)}, pp.\  4489--4495. IEEE, 2018.

\bibitem[Yan et~al.(2020)Yan, Song, Li, Zou, and Ren]{Yan2020Mixup}
Yan, S., Song, H., Li, N., Zou, L., and Ren, L.
\newblock Improve unsupervised domain adaptation with mixup training.
\newblock \emph{arXiv preprint arXiv:2001.00677}, 2020.

\bibitem[Yingzhen \& Mandt(2018)Yingzhen and Mandt]{Li2018DSAE}
Yingzhen, L. and Mandt, S.
\newblock Disentangled sequential autoencoder.
\newblock In \emph{International Conference on Machine Learning}, pp.\
  5670--5679. PMLR, 2018.

\bibitem[Zhou et~al.(2021)Zhou, Yang, Qiao, and Xiang]{Zhou2021arXiv}
Zhou, K., Yang, Y., Qiao, Y., and Xiang, T.
\newblock Domain generalization with mixstyle.
\newblock \emph{arXiv preprint arXiv:2104.02008}, 2021.

\end{thebibliography}
\bibliographystyle{icml2022}

\newpage
\appendix
\onecolumn


\section{Proofs}
\label{app:proofs}

\subsection{Derivation of ELBO for Stationary Environments}
\label{app:proof_stationay}
As we introduce two latent variables to account for the two types of distribution shift, the data generating procedure for one domain can be expressed as:
\begin{equation}
\begin{aligned}
    p(\mathbf{x},\mathbf{y}, \mathbf{z}^c, \mathbf{z}^w,\mathbf{z}^v) 
    &= p(\mathbf{z}^c) p(\mathbf{z}^w) p(\mathbf{z}^v) p(\mathbf{x}|\mathbf{z}^c, \mathbf{z}^w) p(\mathbf{y}|\mathbf{z}^c, \mathbf{z}^v) \\
    &= p(\mathbf{z}^w) p(\mathbf{z}^v) \prod_{i=1}^{N} p(z^c_i) p(x_{i}|z^c_{i}, \mathbf{z}^w)p(y_{i}|z^c_{i},\mathbf{z}^v)
\end{aligned}
\end{equation}

We let $(\mathbf{z}^c,\mathbf{z}^w)$ be the latent variables for $\mathbf{x}$, and $\mathbf{z}^v$ be latent variable for $\mathbf{y}$, thus the distribution of these three latent variables can be inferred from the observable datapoints as $p(\mathbf{z}^c|\mathbf{x})$,$p(\mathbf{z}^w|\mathbf{x})$, $p(\mathbf{z}^v|\mathbf{y})$, respectively. The joint distribution of latent variables is:
\begin{equation}
\label{eq:latent_conditonal_dis}
    p(\mathbf{z}^c, \mathbf{z}^w, \mathbf{z}^v|\mathbf{x}, \mathbf{y}) = p(\mathbf{z}^c|\mathbf{x}) p(\mathbf{z}^w|\mathbf{x})p(\mathbf{z}^v|\mathbf{y})
\end{equation}


For our approximating distribution in Eq~\ref{eq:latent_conditonal_dis}, we can choose the form $q(\mathbf{z}^c, \mathbf{z}^w,\mathbf{z}^v|\mathbf{x},\mathbf{y}) = q(\mathbf{z}^c|\mathbf{x}) q(\mathbf{z}^w|\mathbf{x})q(\mathbf{z}^v|\mathbf{y})$. Thus, we can get:
\begin{equation}
    \begin{aligned}
        \mathbb{D}_{KL}(q, p) &= \mathbb{E}_{q}\log\frac{q(\mathbf{z}^c, \mathbf{z}^w,\mathbf{z}^v|\mathbf{x},\mathbf{y})}{p(\mathbf{z}^c, \mathbf{z}^w,\mathbf{z}^v|\mathbf{x},\mathbf{y})} \\
        &= \mathbb{E}_{q}\log\frac{q(\mathbf{z}^c, \mathbf{z}^w,\mathbf{z}^v|\mathbf{x},\mathbf{y})}{p(\mathbf{x},\mathbf{y},\mathbf{z}^c,\mathbf{z}^w,\mathbf{z}^v)} p(\mathbf{x},\mathbf{y})\\
        &= \mathbb{E}_{q}\log\frac{q(\mathbf{z}^c,\mathbf{z}^w,\mathbf{z}^v|\mathbf{x},\mathbf{y})}{p(\mathbf{x},\mathbf{y},\mathbf{z}^c,\mathbf{z}^w,\mathbf{z}^v)} + \mathbb{E}_{q}\log p(\mathbf{x},\mathbf{y}) \\
        &= \mathbb{E}_{q}\log\frac{q(\mathbf{z}^c,\mathbf{z}^w,\mathbf{z}^v|\mathbf{x},\mathbf{y})}{p(\mathbf{x},\mathbf{y},\mathbf{z}^c,\mathbf{z}^w,\mathbf{z}^v)} + \log p(\mathbf{x},\mathbf{y})
    \end{aligned}
\end{equation}

We can get:
\begin{equation}
    \begin{aligned}
        \log p(\mathbf{x},\mathbf{y}) = \mathbb{D}_{KL}(q, p) + \mathbb{E}_{q}\log\frac{p(\mathbf{x},\mathbf{y},\mathbf{z}^c,\mathbf{z}^w,\mathbf{z}^v)}{q(\mathbf{z}^c,\mathbf{z}^w,\mathbf{z}^v|\mathbf{x},\mathbf{y})}
    \end{aligned}
\end{equation}

As $\mathbb{D}_{KL}(q, p) \geq 0 $, the variational lower bound for $\log p(\mathbf{x},\mathbf{y})$ is:
\begin{equation}
\label{eq:sta_vlb}
    \mathcal{L} = \mathbb{E}_{q}\log\frac{p(\mathbf{x},\mathbf{y},\mathbf{z}^c,\mathbf{z}^w,\mathbf{z}^v)}{q(\mathbf{z}^c,\mathbf{z}^w,\mathbf{z}^v|\mathbf{x},\mathbf{y})}.
\end{equation}

This formulation can be reorganized as:
\begin{equation}
    \begin{aligned}
        \mathcal{L} &= \mathbb{E}_{q}\log\frac{p(\mathbf{x},\mathbf{y}|\mathbf{z}^c,\mathbf{z}^w,\mathbf{z}^v)p(\mathbf{z}^c)p(\mathbf{z}^w)p(\mathbf{z}^v)}{q(\mathbf{z}^c|\mathbf{x})q(\mathbf{z}^w|\mathbf{x})q(\mathbf{z}^v|\mathbf{y})} \\
        &= \mathbb{E}_{q}\log p(\mathbf{x},\mathbf{y}|\mathbf{z}^c, \mathbf{z}^w,\mathbf{z}^v) + \mathbb{E}_{q}\log \frac{p(\mathbf{z}^c)}{q(\mathbf{z}^c|\mathbf{x})} + \mathbb{E}_{q}\log \frac{p(\mathbf{z}^w)}{q(\mathbf{z}^w|\mathbf{x})} + \mathbb{E}_{q}\log \frac{p(\mathbf{z}^v)}{q(\mathbf{z}^v|\mathbf{y})} \\
        &= \mathbb{E}_{q}\log p(\mathbf{x},\mathbf{y}|\mathbf{z}^c,\mathbf{z}^w,\mathbf{z}^v) -
        \mathbb{D}_{KL}(q(\mathbf{z}^c|\mathbf{x}), p(\mathbf{z}^c)) -
        \mathbb{D}_{KL}(q(\mathbf{z}^w|\mathbf{x}), p(\mathbf{z}^w)) - \mathbb{D}_{KL}(q(\mathbf{z}^v|\mathbf{y}), p(\mathbf{z}^v)).
    \end{aligned}
\end{equation}
As $p(\mathbf{x},\mathbf{y}|\mathbf{z}^c,\mathbf{z}^w,\mathbf{z}^v)=p(\mathbf{x}|\mathbf{z}^c,\mathbf{z}^w)p(\mathbf{y}|\mathbf{z}^c,\mathbf{z}^v)$, the above formulation can be rewrote as:
\begin{equation}
    \begin{aligned}
        \mathcal{L} &\geq  \mathbb{E}_{q}\log p(\mathbf{x}|\mathbf{z}^c, \mathbf{z}^w)p(\mathbf{y}|\mathbf{z}^c,\mathbf{z}^v) - \mathbb{D}_{KL}(q(\mathbf{z}^c|\mathbf{x}), p(\mathbf{z}^c)) - 
        \mathbb{D}_{KL}(q(\mathbf{z}^w|\mathbf{x}), p(\mathbf{z}^w)) -
        \mathbb{D}_{KL}(q(\mathbf{z}^v|\mathbf{y}), p(\mathbf{z}^v)). 
    \end{aligned}
\end{equation}

\subsection{Derivation of ELBO for Non-stationary Environments}
\label{app:proof_nonstationay}
We assume the prior distribution of latent variables $\mathbf{z}^w$ and $\mathbf{z}^v$ satisfy Markov property. This means each of them relies on the value of their previous states:
\begin{equation}
    p(\mathbf{z}^w)=p(\mathbf{z}^w_t|\mathbf{z}^w_{<t}), p(\mathbf{z}^v)=p(\mathbf{z}^v_t|\mathbf{z}^v_{<t}),
\end{equation}


The joint distribution of data and latent variables is:
\begin{equation}
\label{eq:dy_joint}
    \begin{aligned}
        & p(\mathbf{x}_{1:T},\mathbf{y}_{1:T},\mathbf{z}^c,\mathbf{z}^w_{1:T},\mathbf{z}^v_{1:T}) \\
        &=  \prod_{t=1}^T p(\mathbf{x}_t, \mathbf{y}_t|\mathbf{z}^c,\mathbf{z}^w_t,\mathbf{z}^v_t) p(\mathbf{z}^c) p(\mathbf{z}^w_t|\mathbf{z}^w_{<t}) p(\mathbf{z}^v_t|\mathbf{z}^v_{<t}) \\
        &= \prod_{t=1}^T p(\mathbf{z}^w_t|\mathbf{z}^w_{<t}) p(\mathbf{z}^v_t|\mathbf{z}^v_{<t}) \prod_{i=1}^{N_t} p(z^c_i) p(x_{i}|z^c_{i}, \mathbf{z}^w_t)p(y_{i}|z^c_{i},\mathbf{z}^v_t) \\
    \end{aligned}
\end{equation}
where $p(\mathbf{z}^w_1)=p(\mathbf{z}^w_1|\mathbf{z}^w_0)$ and $p(\mathbf{z}^v_1)=p(\mathbf{z}^v_1|\mathbf{z}^v_0)$. 



For a non-stationary environment, we assume that $(\mathbf{z}^c,\mathbf{z}^w_t)$ are the latent variables for $\mathbf{x}_t$ and $\mathbf{z}^v_t$ is the latent variable for $\mathbf{y}_t$ at $t$-th time stamp. Thus the distribution of the latent variables for $t$-th time stamp can be express as $p(\mathbf{z}^c|\mathbf{x}_t)$, $p(\mathbf{z}^w_t|\mathbf{x}_t)$ and $p(\mathbf{z}^v_t|\mathbf{y}_t)$, respectively. We employ
$q(\mathbf{z}^c|\mathbf{x}_t)$, $q(\mathbf{z}^w_t|\mathbf{z}^w_{<t},\mathbf{x}_t)$ and $q(\mathbf{z}^v_t|\mathbf{z}^v_{<t},\mathbf{y}_t)$ to approximate the prior distributions here. Similar to Eq~\ref{eq:sta_vlb}, we can draw the variational lower bound for $\log p(\mathbf{x}_{1:T},\mathbf{y}_{1:T})$ as:
\begin{equation}
\label{eq:dy_vlb}
    \mathcal{L} = \mathbb{E}_{q}\log\frac{p(\mathbf{x}_{1:T},\mathbf{y}_{1:T},\mathbf{z}^c,\mathbf{z}^w_{1:T},\mathbf{z}^v_{1:T})}{q(\mathbf{z}^c,\mathbf{z}^w_{1:T},\mathbf{z}^v_{1:T}|\mathbf{x}_{1:T},\mathbf{y}_{1:T})}.
\end{equation}

When incorporating with Eq~\ref{eq:dy_joint}, this can be reorganized as:
\begin{equation}\footnotesize
\begin{aligned}
\mathcal{L} &= \mathbb{E}_{q}\log\frac{\prod_{t=1}^T p(\mathbf{x}_t, \mathbf{y}_t|\mathbf{z}^c,\mathbf{z}^w_t,\mathbf{z}^v_t) p(\mathbf{z}^c) p(\mathbf{z}^w_t|\mathbf{z}^w_{<t}) p(\mathbf{z}^v_t|\mathbf{z}^v_{<t})} {\prod_{t=1}^T q(\mathbf{z}^c|\mathbf{x}_t) q(\mathbf{z}^w_t|\mathbf{z}^w_{<t},\mathbf{x}_t) q(\mathbf{z}^v_t|\mathbf{z}^v_{<t},\mathbf{y}_t)} \\
&=  \mathbb{E}_{q} \Big[ \log\frac{\prod_{t=1}^T p(\mathbf{z}^w_t|\mathbf{z}^w_{<t})}{\prod_{t=1}^T q(\mathbf{z}^w_t|\mathbf{z}^w_{<t},\mathbf{x}_t)} + \log\frac{\prod_{t=1}^T p(\mathbf{z}^v_t|\mathbf{z}^v_{<t})}{\prod_{t=1}^T q(\mathbf{z}^v_t|\mathbf{z}^v_{<t},\mathbf{y}_t)} + 
\log\frac{\prod_{t=1}^T p(\mathbf{z}^c)}{\prod_{t=1}^T q(\mathbf{z}^c|\mathbf{x}_t)} \\
&+\log\prod_{t=1}^T p(\mathbf{x}_t|\mathbf{z}^c_t,\mathbf{z}^w_t)p(\mathbf{y}_t|\mathbf{z}^c_t,\mathbf{z}^v_t) \Big] \\
&= \mathbb{E}_{q} \Big[ - \sum_{t=1}^T\log\frac{q(\mathbf{z}^w_t|\mathbf{z}^w_{<t},\mathbf{x}_t)}{p(\mathbf{z}^w_t|\mathbf{z}^w_{<t})} - \sum_{t=1}^T\log\frac{q(\mathbf{z}^v_t|\mathbf{z}^v_{<t},\mathbf{y}_t)}{p(\mathbf{z}^v_t|\mathbf{z}^v_{<t})} - \sum_{t=1}^T\log \frac{q(\mathbf{z}^c_t|\mathbf{x}_t)}{p(\mathbf{z}^c_t)} \\
&+ \sum_{t=1}^T\log p(\mathbf{x}_t|\mathbf{z}^c,\mathbf{z}^w_t)p(\mathbf{y}_t|\mathbf{z}^c,\mathbf{z}^v_t) \Big]\\
\end{aligned}
\end{equation}


By Jensen’s inequality, the above formulation can be rewrote as:
\begin{equation}
\begin{aligned}
    \mathcal{L} &\geq \mathbb{E}_{q} \Big[ \sum_{t=1}^T\log p(\mathbf{x}_t|\mathbf{z}^c,\mathbf{z}^w_t)p(\mathbf{y}_t|\mathbf{z}^c,\mathbf{z}^v_t) -
    \mathbb{D}_{KL}(q(\mathbf{z}^c|\mathbf{x}_t), p(\mathbf{z}^c)) - 
    \mathbb{D}_{KL}(q(\mathbf{z}^w_t|\mathbf{z}^w_{<t},\mathbf{x}_t), p(\mathbf{z}^w_t|\mathbf{z}^w_{<t})) \\
    &- \mathbb{D}_{KL}(q(\mathbf{z}^v_t|\mathbf{z}^v_{<t},\mathbf{y}_t), p(\mathbf{z}^v_t|\mathbf{z}^v_{<t}))\Big]
\end{aligned}
\end{equation}


This complements the proof.




\section{Probabilistic Generative Model of LSSAE}
\begin{figure*}[!tbp]
    \begin{center}
    \includegraphics[width=0.80\textwidth]{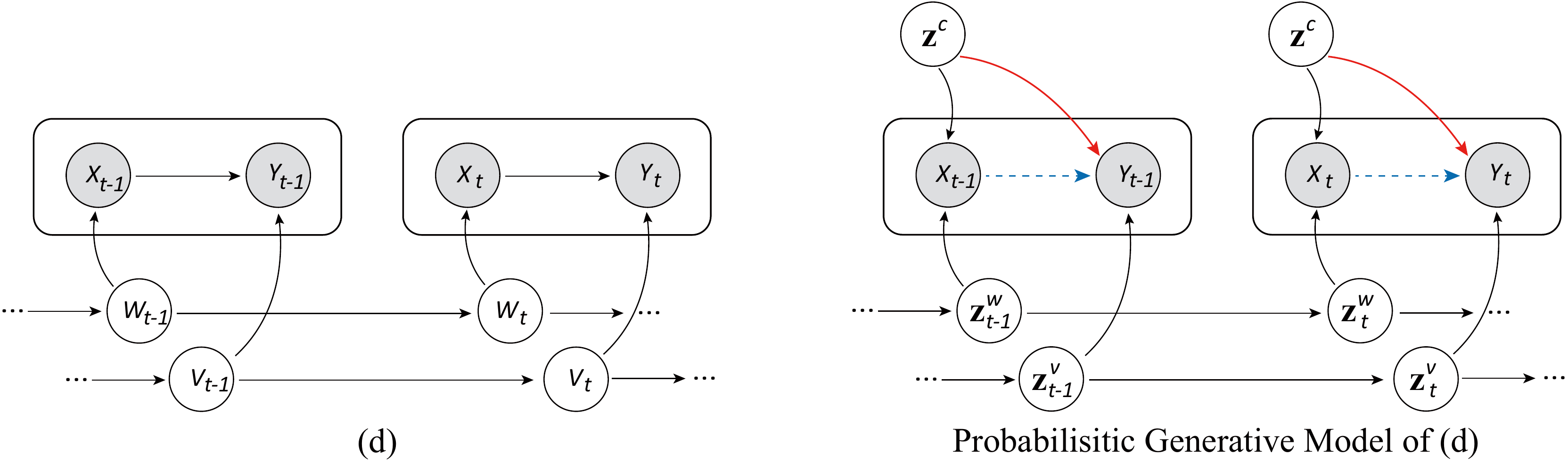}
    \end{center}
    \vspace{-0.3cm}
    \caption{The probabilistic generative graph of LSSAE with $\mathbf{z}^c$ presented. Here, the dependence of $X_t$ on $Y_t$ in Fig.~\ref{fig:DAG}(d) is substituted with $\mathbf{z}^c$ which mainly captures the static category information in data sample space.} 
    \label{fig:PGM}
\end{figure*}

In Fig.~\ref{fig:PGM}, we present the complete probabilistic generative graph of our LSSAE. The left part is the DAG of LSSAE presented in Fig.~\ref{fig:DAG}(d) where we did not include $\mathbf{z}^c$ in this figure as the main focus of Fig.~\ref{fig:DAG} is the dynamic factors (\ie~$W$ and $V$) for presenting our main idea of evolving dynamics at a high level instead of DG. In the right part, we illustrate the relationship between our DAG and probabilistic generative model (with $\mathbf{z}^c$). Here, the dependence of $X_t$ on $Y_t$ (the dotted blue line) is substituted with $\mathbf{z}^c$ (the solid red line) which mainly captures the static category information in data sample space.

\section{Additional Details on the Experimental Setup}
\label{app:additional_setup}

\begin{table*}[t]
\caption{Brief description of employed benchmarks in this work.}
\label{tab:datasets}
\vskip 0.15in
\begin{center}
\begin{tabular}{lcccccc}
\toprule
Dataset & Type & Number & Source Domains & Intermediate Domains & Target Domains & Total Domains\\
\midrule
Circle/-C   & Digital & 3,000  & 15 & 5 & 10 & 30 \\
Sine/-C     & Digital & 2,280  & 12 & 4 & 8  & 24 \\
RMNIST      & Image   & 70,000 & 10 & 3 & 6  & 19 \\
Portraits   & Image   & 37,921 & 19 & 5 & 10 & 34 \\
CalTran     & Image   & 5,450  & 19 & 5 & 10 & 34 \\
PowerSupply & Digital & 29,928 & 15 & 5 & 10 & 30 \\
\bottomrule
\end{tabular}
\end{center}
\end{table*}

\subsection{Datasets}
Our experiments are conducted on 2 synthetic and 4 real-world datasets presented in Table~\ref{tab:datasets}. More detailed description are given below.
\begin{itemize}
    \item \textbf{Circle}~\cite{Pesaranghader2016FastHD}:~Each data in this dataset owns two attributes $(x, y), x, y \in [0, 1]$. The label is assigned using a circle curve as the decision boundary following $(x-x_0)^2+(y-y_0)^2 \leq r^2$, where $(x_0, y_0)$ are the location of the center and $r$ is the radius of this circle. To generate Circle-C, we inject gradual shift via modifying the value of $x_0$ continuously throughout the domains. 
    
    \item \textbf{Sine}~\cite{Pesaranghader2016FastHD}:~Each data in this dataset owns two attributes $(x, y), x, y \in [0, 1]$. The label is assigned using a sine curve as the decision boundary following $y \leq \sin (x)$. To simulate abrupt concept drift for {Sine-C}, labels are reversed (i.e., from 0 to 1 or from 1 to 0) from the $6$-th domain to the last one.
    
    \item \textbf{RMNIST}~\cite{Ghifary2015RMNIST}:~This dataset is composed of MNIST digits of various rotations. We generate 19 domains via applying the rotations with degree of $\mathcal{R}=\{0^\circ,15^\circ,30^\circ,...,180^\circ\}$ on each domain. Note that each image is seen at exactly one angle, so the training procedure cannot track a single image across different angles.
    
    \item \textbf{Portraits}~\cite{Ginosar2015ICCVW}:~This is a real-world dataset of photos collected in American high school seniors. The portraits are taken over 108 years (1905-2013) across 26 states. The goal is to classify the gender for each photo. We split the dataset into $34$ domains by a fixed internal along time.
    
    \item \textbf{CalTran}~\cite{Hoffman2014CVPR}:~This dataset contains real-world images captured by a fixed traffic camera deployed in an intersection over time. Frames were updated at $3$ minute intervals each with a resolution $320 \times 320$. We divide it into $34$ domains by time. This is a scene classification task to determine whether one or more cars are present in, or approaching the intersection. The challenge mainly raise from the continually evolving domain shift as changes include time, illumination, weather, \etc
    
    \item \textbf{PowerSupply}~\cite{Dau2019UCR}:~This dataset is comprised of records of hourly power supply collected by an Italy electricity company. We form $30$ domains according to days. Each data is assigned by a binary class label which represents which time of day the current power supply belongs to (\ie~am or pm). The concept shift may results from the change in season, weather, price or the differences between working days and weekend. 
\end{itemize}

\subsection{Model Architecture \& Hyperparameters}
\label{app:architecture}
Neural network architectures used for each dataset:

\begin{table*}[h]
\caption{Model architectures for different datasets.}
\label{tab:digit_architecture}
\vskip 0.15in
\begin{center}
\begin{tabular}{p{80pt}<{\raggedright} p{85pt}<{\raggedright} p{95pt}<{\raggedright}}
\toprule
\textbf{Dataset} & \textbf{Encoder} & \textbf{Decoder} \\
\midrule
Circle/-C   & \multirow{3}*{Non-linear Encoder} &\multirow{3}*{Non-linear Decoder} \\
Sine/-C       \\
PowerSupply   \\
\hline
RMNIST      & MNIST ConvNet  & MNIST ConvTranNet \\
\hline
Portraits   & \multirow{2}*{ResNet-18}  & \multirow{2}*{ConvTranNet}  \\
CalTran         \\
\bottomrule
\end{tabular}
\end{center}
\end{table*}

\newpage
Neural network architecture for digital experiments~(Circle/-C, Sine/-C, PowerSupply):
\begin{table*}[h]
\caption{Implementation of Non-linear Encoder.}
\label{tab:digit_encoder}
\vskip 0.15in
\begin{center}
\begin{tabular}{p{30pt}<{\raggedright} p{125pt}<{\raggedright}}
\toprule
\textbf{\#} & \textbf{Layer} \\
\midrule
1  & Linear(in=$d$, output=512) \\
2  & ReLU \\
3  & Linear(in=512, output=512) \\
4  & ReLU \\
5  & Linear(in=512, output=512) \\
6  & ReLU \\
7  & Linear(in=512, output=512) \\
\bottomrule
\end{tabular}
\end{center}
\end{table*}

\begin{table*}[h]
\caption{Implementation of Non-linear Decoder.}
\label{tab:digit_decoder}
\vskip 0.15in
\begin{center}
\begin{tabular}{p{30pt}<{\raggedright} p{125pt}<{\raggedright}}
\toprule
\textbf{\#} & \textbf{Layer} \\
\midrule
1  & Linear(in=$d$, output=16) \\
2  & BatchNorm \\
3  & LeakyReLU(0.2) \\
4  & Linear(in=16, output=64) \\
5  & BatchNorm \\
6  & LeakyReLU(0.2) \\
7  & Linear(in=64, output=128) \\
8. & BatchNorm \\
9  & LeakyReLU(0.2) \\
10  & Linear(in=128, output=$d$) \\
\bottomrule
\end{tabular}
\end{center}
\end{table*}

\newpage
Neural network architecture for RMNIST experiments:
\begin{table*}[h]
\caption{Implementation of MNIST ConvNet.}
\label{tab:mnist_encoder}
\vskip 0.15in
\begin{center}
\begin{tabular}{p{30pt}<{\raggedright} p{185pt}<{\raggedright}}
\toprule
\textbf{\#} & \textbf{Layer} \\
\midrule
1  & Conv2D(in=$d$, output=64) \\
2  & ReLU \\
3  & GroupNorm(groupds=8) \\
4  & Conv2D(in=64, output=128, stride=2) \\
5  & ReLU \\
6  & GroupNorm(groupds=8) \\
7  & Conv2D(in=128, output=128) \\
8  & ReLU \\
9  & GroupNorm(groupds=8) \\
10  & Conv2D(in=128, output=128) \\
11  & ReLU \\
12  & GroupNorm(groupds=8) \\
13  & Global average-pooling \\
\bottomrule
\end{tabular}
\end{center}
\end{table*}

\begin{table*}[h]
\caption{Implementation of MNIST ConvTranNet.}
\label{tab:mnist_decoder}
\vskip 0.15in
\begin{center}
\begin{tabular}{p{30pt}<{\raggedright} p{210pt}<{\raggedright}}
\toprule
\textbf{\#} & \textbf{Layer} \\
\midrule
1  & Linear(in=$d$, output=1024) \\
2  & BatchNorm \\
3  & ReLU \\
4  & Upsample(8) \\
5  & ConvTransposed2D(in=64, output=128, kernel=5) \\
6  & BatchNorm \\
7  & ReLU \\
8  & Upsample(24) \\
9  & ConvTransposed2D(in=128, output=256, kernel=5) \\
10  & BatchNorm \\
11  & ReLU \\
12 & Conv2D(in=256, output=1, kernel=1) \\
13 & Sigmoid \\
\bottomrule
\end{tabular}
\end{center}
\end{table*}

\newpage
Neural network architecture for Portraits and CalTran experiments:
\begin{table*}[h]
\caption{Implementation of ConvTranNet.}
\label{tab:nature_decoder}
\vskip 0.15in
\begin{center}
\begin{tabular}{p{30pt}<{\raggedright} p{210pt}<{\raggedright}}
\toprule
\textbf{\#} & \textbf{Layer} \\
\midrule
1  & Linear(in=$d$, output=1024) \\
2  & BatchNorm \\
3  & ReLU \\
4  & Upsample(16) \\
5  & ConvTransposed2D(in=64, output=128, kernel=5) \\
6  & BatchNorm \\
7  & ReLU \\
8  & Upsample(40) \\
9  & ConvTransposed2D(in=128, output=256, kernel=5) \\
10 & BatchNorm \\
11 & ReLU \\
12 & Upsample(80) \\
13 & ConvTransposed2D(in=256, output=3, kernel=5) \\
14 & BatchNorm \\
15 & ReLU \\
16 & Sigmoid \\
\bottomrule
\end{tabular}
\end{center}
\end{table*}

The model architecture of the encoder is ResNet-18, we replace the final softmax layer of the official version following~\citet{Gulrajani2021ICLR}. Besides, a dropout layer before the final linear layer is inserted. This network is initialized by random rather than loading the pretrained parameters on ImageNet.

\newpage
We list the values of hyperparameters for different datasets below. All models are optimized by Adam~\cite{Kingma2015Adam}. In our experiments, we found that keeping the balance of the three KL divergence terms for $\mathbf{z}^c$, $\mathbf{z}^w$ and $\mathbf{z}^v$ via adjusting the value of $\lambda_1$, $\lambda_2$ and $\lambda_3$ is beneficial for the final results. 

\begin{table*}[!h]
\caption{Hyperparametes and their default values.}
\label{tab:hyperparameter}
\vskip 0.15in
\begin{center}
\begin{tabular}{p{80pt}<{\raggedright} p{200pt}<{\raggedright} p{80pt}<{\raggedright}}
\toprule
\textbf{Dataset} & \textbf{Parameters} & \textbf{Value} \\
\midrule
\multirow{5}*{Circle} & learning rate for $E^c, D, C$ & 5e-5\\
& learning rate for $E^w, F^w, E^v, F^v$ & 5e-6 \\
& batch size & 24 \\
& $\lambda_1, \lambda_2, \lambda_3$ & 1.0, 1.0, 1.0 \\
& $\alpha$ & 0.05 \\
\hline
\multirow{5}*{Circle-C} & learning rate for $E^c, D, C$ & 1e-5\\
& learning rate for $E^w, F^w, E^v, F^v$ & 1e-6 \\
& batch size & 24 \\
& $\lambda_1, \lambda_2, \lambda_3$ & 1.0, 2.0, 1.0 \\
& $\alpha$ & 0.05 \\
\hline
\multirow{5}*{Sine} & learning rate for $E^c, D, C$ & 5e-5\\
& learning rate for $E^w, F^w, E^v, F^v$ & 5e-6 \\
& batch size & 24 \\
& $\lambda_1, \lambda_2, \lambda_3$ & 2.0, 1.0, 1.0 \\
& $\alpha$ & 0.05 \\
\hline
\multirow{5}*{Sine-C} & learning rate for $E^c, D, C$ & 1e-5\\
& learning rate for $E^w, F^w, E^v, F^v$ & 1e-6 \\
& batch size & 24 \\
& $\lambda_1, \lambda_2, \lambda_3$ & 2.0, 1.0, 1.0 \\
& $\alpha$ & 0.05 \\
\hline
\multirow{5}*{RMNIST} & learning rate for $E^c, D, C$ & 1e-3\\
& learning rate for $E^w, F^w, E^v, F^v$ & 1e-4 \\
& batch size & 48 \\
& $\lambda_1, \lambda_2, \lambda_3$ & 2.0, 1.0, 1.0 \\
& $\alpha$ & 0.05 \\
\hline
\multirow{5}*{Portraits} & learning rate for $E^c, D, C$ & 1e-5\\
& learning rate for $E^w, F^w, E^v, F^v$ & 1e-6 \\
& batch size & 24 \\
& $\lambda_1, \lambda_2, \lambda_3$ & 0.5, 1.0, 1.0 \\
& $\alpha$ & 0.05 \\
\hline
\multirow{5}*{CalTran} & learning rate for $E^c, D, C$ & 5e-5\\
& learning rate for $E^w, F^w, E^v, F^v$ & 5e-6 \\
& batch size & 24 \\
& $\lambda_1, \lambda_2, \lambda_3$ & 1.0, 1.0, 1.0 \\
& $\alpha$ & 0.1 \\
\hline
\multirow{5}*{PowerSupply} & learning rate for $E^c, D, C$ & 1e-5\\
& learning rate for $E^w, F^w, E^v, F^v$ & 1e-6 \\
& batch size & 48 \\
& $\lambda_1, \lambda_2, \lambda_3$ & 2.0, 1.0, 1.0 \\
& $\alpha$ & 0.1 \\
\bottomrule
\end{tabular}
\end{center}
\end{table*}

\newpage
\section{Additional Experimental Results}
In this section, we provide more experimental results for our proposed evolving domain generalization task. As we can see, in most of the cases, we can achieve the state-of-the-art performance compared with other domain generalization baselines. However, in some cases, our proposed method cannot achieve desired results. For example, in Sine-C, ERM and some domain generalization baselines can achieve a desired classification performance especially for domain index 17,18 and 19, which we conjecture the reason that their decision boundaries overfit to the abrupt concept shift, which further leads to their poor performance on the following domains after domain 19.  However, our proposed method aims to learn the evolving pattern starting from $t=0$, which may not be optimal to suit to this abrupt concept shift case. One possible future direction is only to learn the evolving pattern in a certain time duration (i.e., time duration in $[t-T_0,t]$, where $t$ is the current time stamp). Nevertheless, our proposed method shows its effectiveness in dynamic modeling for evolving domain generalization task. We will leave the discussion how to find a suitable time duration (i.e., a suitable $T_0$) in our future work.   
\label{app:additional_exp}
\begin{table*}[h]
\caption{Circle. We show the results on each target domain by domain index.}
\vskip 0.15in
\begin{center}
\begin{scriptsize}
\begin{tabular}{cccccccccccc}
\toprule
Algorithm & \textbf{21} & \textbf{22} & \textbf{23} & \textbf{24} & \textbf{25} & \textbf{26} & \textbf{27} & \textbf{28} & \textbf{29} & \textbf{30} & \textcolor{teal}{Avg}\\
\midrule
ERM     & 53.9 $\pm$ 3.5 & 55.8 $\pm$ 4.8 & 53.9 $\pm$ 5.2 & 44.7 $\pm$ 6.3 & 56.9 $\pm$ 4.4 & 47.8 $\pm$ 5.8 & 41.9 $\pm$ 7.5 & 41.7 $\pm$ 5.9 & 54.2 $\pm$ 3.0 & 47.8 $\pm$ 5.7 & 49.9  \\
Mixup   & 48.6 $\pm$ 3.8 & 51.7 $\pm$ 4.0 & 49.4 $\pm$ 4.5 & 43.6 $\pm$ 5.8 & 56.9 $\pm$ 4.4 & 47.8 $\pm$ 5.8 & 41.9 $\pm$ 7.5 & 41.7 $\pm$ 5.9 & 54.2 $\pm$ 3.0 & 47.8 $\pm$ 5.7 & 48.4  \\
MMD     & 50.0 $\pm$ 3.9 & 53.6 $\pm$ 4.4 & 55.0 $\pm$ 4.3 & 51.9 $\pm$ 6.0 & 60.8 $\pm$ 3.9 & 49.7 $\pm$ 7.2 & 41.9 $\pm$ 7.5 & 41.7 $\pm$ 5.9 & 54.2 $\pm$ 3.0 & 47.8 $\pm$ 5.7 & 50.7  \\
MLDG    & 57.8 $\pm$ 3.6 & 57.7 $\pm$ 5.0 & 55.3 $\pm$ 4.9 & 46.4 $\pm$ 6.8 & 56.9 $\pm$ 4.4 & 47.8 $\pm$ 5.8 & 41.9 $\pm$ 7.5 & 41.7 $\pm$ 5.9 & 54.2 $\pm$ 3.0 & 47.8 $\pm$ 5.7 & 50.8  \\
IRM     & 57.8 $\pm$ 3.9 & 59.4 $\pm$ 5.4 & 56.9 $\pm$ 4.9 & 48.1 $\pm$ 7.4 & 57.5 $\pm$ 4.3 & 47.8 $\pm$ 5.8 & 41.9 $\pm$ 7.5 & 41.7 $\pm$ 5.9 & 54.2 $\pm$ 3.0 & 47.8 $\pm$ 5.7 & 51.3  \\
RSC     & 45.3 $\pm$ 3.6 & 51.4 $\pm$ 3.8 & 49.4 $\pm$ 4.5 & 43.6 $\pm$ 5.8 & 56.9 $\pm$ 4.4 & 47.8 $\pm$ 5.8 & 41.9 $\pm$ 7.5 & 41.7 $\pm$ 5.9 & 54.2 $\pm$ 3.0 & 47.8 $\pm$ 5.7 & 48.0  \\
MTL     & 61.4 $\pm$ 2.2 & 57.2 $\pm$ 6.4 & 53.3 $\pm$ 5.1 & 48.3 $\pm$ 6.2 & 56.9 $\pm$ 4.8 & 49.2 $\pm$ 3.7 & 43.3 $\pm$ 5.0 & 45.8 $\pm$ 2.8 & 54.2 $\pm$ 5.7 & 42.2 $\pm$ 4.9 & 51.2  \\
Fish    & 51.7 $\pm$ 3.7 & 53.1 $\pm$ 3.7 & 49.4 $\pm$ 4.5 & 43.6 $\pm$ 5.8 & 56.9 $\pm$ 4.4 & 47.8 $\pm$ 5.8 & 41.9 $\pm$ 7.5 & 41.7 $\pm$ 5.9 & 54.2 $\pm$ 3.0 & 47.8 $\pm$ 5.7 & 48.8  \\
CORAL   & 65.3 $\pm$ 3.2 & 63.9 $\pm$ 4.4 & 60.0 $\pm$ 4.8 & 56.4 $\pm$ 6.0 & 60.2 $\pm$ 4.3 & 47.8 $\pm$ 5.8 & 41.9 $\pm$ 7.5 & 41.7 $\pm$ 5.9 & 54.2 $\pm$ 3.0 & 47.8 $\pm$ 5.7 & 53.9  \\
AndMask & 42.8 $\pm$ 3.4 & 50.6 $\pm$ 4.1 & 49.4 $\pm$ 4.5 & 43.6 $\pm$ 5.8 & 56.9 $\pm$ 4.4 & 47.8 $\pm$ 5.8 & 41.9 $\pm$ 7.5 & 41.7 $\pm$ 5.9 & 54.2 $\pm$ 3.0 & 49.7 $\pm$ 5.4 & 47.9  \\
DIVA    & 81.3 $\pm$ 3.5 & 76.3 $\pm$ 4.2 & 74.7 $\pm$ 4.6 & 56.7 $\pm$ 5.1 & 67.0 $\pm$ 6.1 & 62.3 $\pm$ 5.1 & 62.0 $\pm$ 5.6 & 66.3 $\pm$ 4.1 & 70.3 $\pm$ 5.6 & 62.0 $\pm$ 4.2 & 67.9  \\
LSSAE~(Ours) & 95.8 $\pm$ 1.9 & 95.6 $\pm$ 2.1 & 93.5 $\pm$ 2.9 & 96.3 $\pm$ 1.8 & 83.8 $\pm$ 5.2 & 74.3 $\pm$ 3.6 & 51.9 $\pm$ 5.6 & 52.3 $\pm$ 8.1 & 46.5 $\pm$ 9.2 & 48.4 $\pm$ 5.3 & 73.8  \\
\bottomrule
\end{tabular}
\end{scriptsize}
\end{center}
\vskip -0.1in
\end{table*}


\begin{table*}[h]
\caption{Sine. We show the results on each target domain denoted by domain index.}
\vskip 0.15in
\begin{center}
\begin{small}
\begin{tabular}{cccccccccc}
\toprule
Algorithm & \textbf{17} & \textbf{18} & \textbf{19} & \textbf{20} & \textbf{21} & \textbf{22} & \textbf{23} & \textbf{24} & \textcolor{teal}{Avg}\\
\midrule

ERM     & 71.4 $\pm$ 6.1 & 91.0 $\pm$ 1.5 & 81.6 $\pm$ 2.4 & 53.4 $\pm$ 2.9 & 51.1 $\pm$ 6.7 & 54.3 $\pm$ 4.7 & 49.5 $\pm$ 4.8 & 51.7 $\pm$ 5.0 & 63.0  \\
Mixup   & 63.1 $\pm$ 5.9 & 93.5 $\pm$ 1.7 & 80.6 $\pm$ 3.8 & 52.8 $\pm$ 2.9 & 60.3 $\pm$ 7.2 & 54.2 $\pm$ 2.7 & 49.5 $\pm$ 4.4 & 49.3 $\pm$ 8.0 & 62.9  \\
MMD     & 57.0 $\pm$ 4.2 & 57.1 $\pm$ 4.1 & 47.6 $\pm$ 5.4 & 50.0 $\pm$ 1.8 & 55.1 $\pm$ 6.7 & 54.4 $\pm$ 4.7 & 49.5 $\pm$ 4.8 & 51.7 $\pm$ 5.0 & 55.8  \\
MLDG    & 69.2 $\pm$ 4.2 & 67.7 $\pm$ 4.1 & 52.1 $\pm$ 5.4 & 50.7 $\pm$ 1.8 & 51.1 $\pm$ 6.7 & 54.3 $\pm$ 4.7 & 49.5 $\pm$ 4.8 & 51.7 $\pm$ 5.0 & 63.2  \\
IRM     & 66.9 $\pm$ 6.2 & 81.1 $\pm$ 3.2 & 88.5 $\pm$ 3.0 & 56.6 $\pm$ 6.0 & 57.2 $\pm$ 5.8 & 53.7 $\pm$ 5.1 & 49.5 $\pm$ 2.2 & 51.7 $\pm$ 5.4 & 63.2  \\
RSC     & 61.3 $\pm$ 6.6 & 83.5 $\pm$ 1.9 & 84.5 $\pm$ 2.6 & 52.8 $\pm$ 2.8 & 55.1 $\pm$ 6.7 & 54.4 $\pm$ 4.7 & 49.5 $\pm$ 4.8 & 51.7 $\pm$ 5.0 & 61.5  \\
MTL     & 70.6 $\pm$ 6.6 & 91.6 $\pm$ 1.2 & 79.9 $\pm$ 3.4 & 51.0 $\pm$ 4.7 & 60.3 $\pm$ 7.6 & 53.6 $\pm$ 5.2 & 49.5 $\pm$ 5.3 & 46.9 $\pm$ 5.9 & 62.9  \\
Fish    & 66.1 $\pm$ 6.9 & 82.0 $\pm$ 2.7 & 87.5 $\pm$ 2.4 & 55.2 $\pm$ 3.0 & 51.1 $\pm$ 6.7 & 54.3 $\pm$ 4.7 & 49.5 $\pm$ 4.8 & 51.7 $\pm$ 5.0 & 62.3  \\
CORAL   & 60.0 $\pm$ 5.3 & 57.1 $\pm$ 4.2 & 48.6 $\pm$ 6.4 & 50.7 $\pm$ 1.8 & 49.7 $\pm$ 6.2 & 48.6 $\pm$ 4.6 & 46.3 $\pm$ 5.0 & 51.7 $\pm$ 5.0 & 51.6  \\
AndMask & 44.2 $\pm$ 5.1 & 42.9 $\pm$ 4.2 & 54.2 $\pm$ 7.0 & 71.9 $\pm$ 1.9 & 86.4 $\pm$ 3.2 & 90.4 $\pm$ 2.9 & 88.1 $\pm$ 3.4 & 76.4 $\pm$ 3.7 & 69.3  \\
DIVA    & 79.0 $\pm$ 6.6 & 60.8 $\pm$ 1.9 & 47.6 $\pm$ 2.6 & 50.0 $\pm$ 2.8 & 55.1 $\pm$ 6.7 & 51.9 $\pm$ 4.7 & 38.6 $\pm$ 4.8 & 40.4 $\pm$ 5.0 & 52.9  \\
LSSAE~(Ours) & 93.0 $\pm$ 1.7 & 86.9 $\pm$ 0.7 & 69.2 $\pm$ 1.5 & 63.8 $\pm$ 3.8 & 68.8 $\pm$ 2.5 & 76.8 $\pm$ 4.8 & 63.9 $\pm$ 1.3 & 49.0 $\pm$ 3.1 & 71.4  \\

\bottomrule
\end{tabular}
\end{small}
\end{center}
\vskip -0.1in
\end{table*}


\begin{table*}[h]
\caption{RMNIST. We show the results on each target domain denoted by rotation angle.}
\vskip 0.15in
\begin{center}
\begin{small}
\begin{tabular}{cccccccc}
\toprule
Algorithm & $\bm{130^\circ}$ & $\bm{140^\circ}$ & $\bm{150^\circ}$ & $\bm{160^\circ}$ & $\bm{170^\circ}$ & $\bm{180^\circ}$ & \textcolor{teal}{Avg}\\
\midrule
ERM     & 56.8 $\pm$ 0.9 & 44.2 $\pm$ 0.8 & 37.8 $\pm$ 0.6 & 38.3 $\pm$ 0.8 & 40.9 $\pm$ 0.8 & 43.6 $\pm$ 0.8 & 43.6  \\
Mixup   & 61.3 $\pm$ 0.7 & 47.4 $\pm$ 0.8 & 39.1 $\pm$ 0.7 & 38.3 $\pm$ 0.7 & 40.5 $\pm$ 0.8 & 42.8 $\pm$ 0.9 & 44.9  \\
MMD     & 59.2 $\pm$ 0.9 & 46.0 $\pm$ 0.8 & 39.0 $\pm$ 0.7 & 39.3 $\pm$ 0.8 & 41.6 $\pm$ 0.7 & 43.7 $\pm$ 0.8 & 44.8  \\
MLDG    & 57.4 $\pm$ 0.7 & 44.5 $\pm$ 0.9 & 37.5 $\pm$ 0.8 & 37.5 $\pm$ 0.8 & 39.9 $\pm$ 0.8 & 42.0 $\pm$ 0.9 & 43.1  \\
IRM     & 47.7 $\pm$ 0.9 & 38.5 $\pm$ 0.7 & 34.1 $\pm$ 0.7 & 35.7 $\pm$ 0.8 & 37.8 $\pm$ 0.8 & 40.3 $\pm$ 0.8 & 39.0  \\
RSC     & 54.1 $\pm$ 0.9 & 41.9 $\pm$ 0.8 & 35.8 $\pm$ 0.7 & 37.0 $\pm$ 0.8 & 39.8 $\pm$ 0.8 & 41.6 $\pm$ 0.8 & 41.7  \\
MTL     & 54.8 $\pm$ 0.9 & 43.1 $\pm$ 0.8 & 36.4 $\pm$ 0.8 & 36.1 $\pm$ 0.8 & 39.1 $\pm$ 0.9 & 40.9 $\pm$ 0.8 & 41.7  \\
Fish    & 60.8 $\pm$ 0.8 & 47.8 $\pm$ 0.8 & 39.2 $\pm$ 0.8 & 37.6 $\pm$ 0.7 & 39.0 $\pm$ 0.8 & 40.7 $\pm$ 0.7 & 44.2  \\
CORAL   & 58.8 $\pm$ 0.9 & 46.2 $\pm$ 0.8 & 38.9 $\pm$ 0.7 & 38.5 $\pm$ 0.8 & 41.3 $\pm$ 0.8 & 43.5 $\pm$ 0.8 & 44.5  \\
AndMask & 53.5 $\pm$ 0.9 & 42.9 $\pm$ 0.8 & 37.8 $\pm$ 0.7 & 38.6 $\pm$ 0.8 & 40.8 $\pm$ 0.8 & 43.2 $\pm$ 0.8 & 42.8  \\
DIVA    & 58.3 $\pm$ 0.8 & 45.0 $\pm$ 0.8 & 37.6 $\pm$ 0.8 & 36.9 $\pm$ 0.7 & 38.1 $\pm$ 0.8 & 40.1 $\pm$ 0.8 & 42.7  \\
LSSAE~(Ours) & 64.1 $\pm$ 0.8 & 51.6 $\pm$ 0.8 & 43.4 $\pm$ 0.8 & 38.6 $\pm$ 0.7 & 40.3 $\pm$ 0.8 & 40.4 $\pm$ 0.8 & 46.4 \\

\bottomrule
\end{tabular}
\end{small}
\end{center}
\vskip -0.1in
\end{table*}

\begin{table*}[h]
\caption{Portraits. We show the results on each target domain denoted by domain index.}
\vskip 0.15in
\begin{center}
\begin{scriptsize}
\begin{tabular}{cccccccccccc}
\toprule
Algorithm & \textbf{25} & \textbf{26} & \textbf{27} & \textbf{28} & \textbf{29} & \textbf{30} & \textbf{31} & \textbf{32} & \textbf{33} & \textbf{34} & \textcolor{teal}{Avg}\\
\midrule
ERM     & 75.5 $\pm$ 0.9 & 83.8 $\pm$ 0.9 & 88.5 $\pm$ 0.8 & 93.3 $\pm$ 0.7 & 93.4 $\pm$ 0.6 & 92.1 $\pm$ 0.7 & 90.6 $\pm$ 0.8 & 84.3 $\pm$ 0.9 & 88.5 $\pm$ 0.9 & 87.9 $\pm$ 1.4 & 87.8  \\
Mixup   & 75.5 $\pm$ 0.9 & 83.8 $\pm$ 0.9 & 88.5 $\pm$ 0.8 & 93.3 $\pm$ 0.7 & 93.4 $\pm$ 0.6 & 92.1 $\pm$ 0.7 & 90.6 $\pm$ 0.8 & 84.3 $\pm$ 0.9 & 88.5 $\pm$ 0.9 & 87.9 $\pm$ 1.4 & 87.8  \\
MMD     & 74.0 $\pm$ 1.0 & 83.8 $\pm$ 0.8 & 87.2 $\pm$ 0.8 & 93.0 $\pm$ 0.7 & 93.0 $\pm$ 0.6 & 91.9 $\pm$ 0.7 & 90.9 $\pm$ 0.7 & 84.7 $\pm$ 1.4 & 88.3 $\pm$ 0.9 & 85.8 $\pm$ 1.8 & 87.3  \\
MLDG    & 76.4 $\pm$ 0.8 & 85.5 $\pm$ 0.9 & 90.1 $\pm$ 0.7 & 94.3 $\pm$ 0.6 & 93.5 $\pm$ 0.6 & 92.0 $\pm$ 0.7 & 90.8 $\pm$ 0.8 & 85.6 $\pm$ 1.1 & 89.3 $\pm$ 0.8 & 87.6 $\pm$ 1.6 & 88.5  \\
IRM     & 74.2 $\pm$ 0.9 & 83.5 $\pm$ 0.9 & 88.5 $\pm$ 0.8 & 91.0 $\pm$ 0.8 & 90.4 $\pm$ 0.7 & 87.3 $\pm$ 0.8 & 87.0 $\pm$ 0.9 & 80.4 $\pm$ 1.5 & 86.7 $\pm$ 0.9 & 85.1 $\pm$ 1.8 & 85.4  \\
RSC     & 75.2 $\pm$ 0.9 & 84.7 $\pm$ 0.8 & 87.9 $\pm$ 0.7 & 93.3 $\pm$ 0.7 & 92.5 $\pm$ 0.7 & 91.0 $\pm$ 0.7 & 90.0 $\pm$ 0.7 & 84.6 $\pm$ 1.2 & 88.2 $\pm$ 0.8 & 85.8 $\pm$ 1.9 & 87.3  \\
MTL     & 78.2 $\pm$ 0.9 & 86.5 $\pm$ 0.8 & 90.9 $\pm$ 0.8 & 94.2 $\pm$ 0.7 & 93.8 $\pm$ 0.6 & 92.0 $\pm$ 0.7 & 91.2 $\pm$ 0.7 & 86.0 $\pm$ 1.2 & 89.3 $\pm$ 0.8 & 87.4 $\pm$ 1.4 & 89.0  \\
Fish    & 78.6 $\pm$ 0.9 & 86.9 $\pm$ 0.8 & 89.5 $\pm$ 0.8 & 93.5 $\pm$ 0.7 & 93.3 $\pm$ 0.6 & 92.1 $\pm$ 0.6 & 91.1 $\pm$ 0.7 & 86.2 $\pm$ 1.3 & 88.7 $\pm$ 0.9 & 87.7 $\pm$ 1.6 & 88.8  \\
CORAL   & 74.6 $\pm$ 0.9 & 84.6 $\pm$ 0.8 & 87.9 $\pm$ 0.8 & 93.3 $\pm$ 0.6 & 92.7 $\pm$ 0.7 & 91.5 $\pm$ 0.7 & 90.7 $\pm$ 0.7 & 84.6 $\pm$ 1.5 & 88.1 $\pm$ 0.9 & 85.9 $\pm$ 1.9 & 87.4  \\
AndMask & 62.0 $\pm$ 1.1 & 70.8 $\pm$ 1.1 & 67.0 $\pm$ 1.2 & 70.2 $\pm$ 1.1 & 75.2 $\pm$ 1.1 & 74.1 $\pm$ 1.0 & 72.7 $\pm$ 1.1 & 64.7 $\pm$ 1.6 & 77.3 $\pm$ 1.1 & 74.9 $\pm$ 2.1 & 70.9  \\
DIVA    & 76.2 $\pm$ 1.0 & 86.6 $\pm$ 0.8 & 88.8 $\pm$ 0.8 & 93.5 $\pm$ 0.7 & 93.1 $\pm$ 0.6 & 91.6 $\pm$ 0.6 & 91.1 $\pm$ 0.7 & 84.7 $\pm$ 1.3 & 89.1 $\pm$ 0.8 & 87.0 $\pm$ 1.5 & 88.2  \\
LSSAE~(Ours) & 77.7 $\pm$ 0.9 & 87.1 $\pm$ 0.8 & 90.8 $\pm$ 0.7 & 94.3 $\pm$ 0.6 & 94.3 $\pm$ 0.6 & 92.2 $\pm$ 0.6 & 91.2 $\pm$ 0.7 & 86.7 $\pm$ 1.1 & 89.6 $\pm$ 0.8 & 86.9 $\pm$ 1.4  & 89.1  \\
\bottomrule
\end{tabular}
\end{scriptsize}
\end{center}
\vskip -0.1in
\end{table*}

\begin{table*}[h]
\caption{CalTran. We show the results on each target domain denoted by domain index.}
\vskip 0.15in
\begin{center}
\begin{scriptsize}
\begin{tabular}{cccccccccccc}
\toprule
Algorithm & \textbf{25} & \textbf{26} & \textbf{27} & \textbf{28} & \textbf{29} & \textbf{30} & \textbf{31} & \textbf{32} & \textbf{33} & \textbf{34} & \textcolor{teal}{Avg}\\
\midrule

ERM     & 29.9 $\pm$ 3.5 & 88.4 $\pm$ 2.1 & 61.1 $\pm$ 3.5 & 56.3 $\pm$ 3.2 & 90.0 $\pm$ 1.6 & 60.1 $\pm$ 2.5 & 55.5 $\pm$ 3.5 & 88.8 $\pm$ 2.4 & 57.1 $\pm$ 3.5 & 50.5 $\pm$ 5.2 & 66.3  \\
Mixup   & 53.6 $\pm$ 3.9 & 89.0 $\pm$ 2.0 & 61.8 $\pm$ 2.4 & 55.7 $\pm$ 2.9 & 88.2 $\pm$ 2.1 & 58.6 $\pm$ 3.0 & 52.3 $\pm$ 3.7 & 88.6 $\pm$ 2.7 & 57.1 $\pm$ 3.0 & 55.1 $\pm$ 4.3 & 66.0  \\
MMD     & 30.2 $\pm$ 2.1 & 92.7 $\pm$ 1.7 & 56.4 $\pm$ 3.7 & 39.1 $\pm$ 3.2 & 93.6 $\pm$ 1.7 & 52.1 $\pm$ 3.2 & 42.8 $\pm$ 3.0 & 92.1 $\pm$ 2.2 & 42.1 $\pm$ 3.8 & 29.4 $\pm$ 3.8 & 57.1  \\
MLDG    & 54.8 $\pm$ 4.1 & 88.6 $\pm$ 2.6 & 62.2 $\pm$ 3.6 & 55.1 $\pm$ 4.1 & 88.3 $\pm$ 1.7 & 60.9 $\pm$ 4.3 & 51.7 $\pm$ 2.6 & 89.0 $\pm$ 1.9 & 56.5 $\pm$ 3.4 & 55.3 $\pm$ 4.8 & 66.2  \\
IRM     & 46.4 $\pm$ 3.7 & 90.8 $\pm$ 1.7 & 60.8 $\pm$ 3.4 & 52.9 $\pm$ 3.1 & 91.8 $\pm$ 1.7 & 56.6 $\pm$ 3.1 & 52.1 $\pm$ 2.9 & 90.9 $\pm$ 2.6 & 55.6 $\pm$ 3.9 & 43.1 $\pm$ 5.5 & 64.1  \\
RSC     & 57.2 $\pm$ 3.0 & 88.4 $\pm$ 2.6 & 62.6 $\pm$ 3.0 & 56.5 $\pm$ 3.7 & 88.0 $\pm$ 2.4 & 59.4 $\pm$ 3.0 & 51.9 $\pm$ 2.9 & 90.0 $\pm$ 2.0 & 59.4 $\pm$ 2.9 & 56.0 $\pm$ 3.1 & 67.0  \\
MTL     & 64.2 $\pm$ 3.0 & 87.2 $\pm$ 2.5 & 64.9 $\pm$ 3.9 & 60.0 $\pm$ 4.8 & 84.5 $\pm$ 2.2 & 60.6 $\pm$ 3.5 & 52.6 $\pm$ 3.7 & 83.9 $\pm$ 2.9 & 58.2 $\pm$ 4.1 & 65.7 $\pm$ 5.6 & 68.2  \\
Fish    & 61.1 $\pm$ 3.5 & 88.2 $\pm$ 1.5 & 64.7 $\pm$ 4.0 & 57.9 $\pm$ 3.1 & 88.3 $\pm$ 2.2 & 59.9 $\pm$ 3.0 & 57.5 $\pm$ 2.7 & 87.4 $\pm$ 2.8 & 57.7 $\pm$ 3.7 & 63.0 $\pm$ 6.1 & 68.6  \\
CORAL   & 50.4 $\pm$ 3.0 & 90.8 $\pm$ 2.0 & 61.2 $\pm$ 3.8 & 55.0 $\pm$ 2.5 & 92.0 $\pm$ 1.7 & 56.8 $\pm$ 3.8 & 52.0 $\pm$ 3.8 & 90.9 $\pm$ 1.6 & 56.8 $\pm$ 2.4 & 50.9 $\pm$ 5.6 & 65.7  \\
AndMask & 30.0 $\pm$ 2.2 & 92.7 $\pm$ 1.7 & 56.2 $\pm$ 3.8 & 39.1 $\pm$ 3.2 & 93.6 $\pm$ 1.7 & 51.6 $\pm$ 3.2 & 42.6 $\pm$ 2.9 & 92.1 $\pm$ 2.2 & 41.2 $\pm$ 3.7 & 29.9 $\pm$ 3.6 & 56.9  \\
DIVA    & 60.6 $\pm$ 2.9 & 90.1 $\pm$ 1.7 & 67.5 $\pm$ 3.1 & 58.9 $\pm$ 3.5 & 88.4 $\pm$ 2.8 & 58.7 $\pm$ 3.3 & 53.8 $\pm$ 3.6 & 89.8 $\pm$ 1.7 & 61.8 $\pm$ 4.8 & 62.0 $\pm$ 3.4 & 69.2  \\
LSSVAE & 63.4 $\pm$ 3.4 & 92.1 $\pm$ 2.0 & 62.6 $\pm$ 4.7 & 58.8 $\pm$ 4.4 & 92.9 $\pm$ 1.6 & 62.0 $\pm$ 3.9 & 54.3 $\pm$ 3.0 & 92.1 $\pm$ 2.2 & 60.5 $\pm$ 3.8 & 67.4 $\pm$ 3.6 & 70.6  \\
\bottomrule
\end{tabular}
\end{scriptsize}
\end{center}
\vskip -0.1in
\end{table*}

\begin{table*}[h]
\caption{Circle-C. We show the results on each target domain denoted by domain index.}
\vskip 0.15in
\begin{center}
\begin{scriptsize}
\begin{tabular}{cccccccccccc}
\toprule
Algorithm & \textbf{21} & \textbf{22} & \textbf{23} & \textbf{24} & \textbf{25} & \textbf{26} & \textbf{27} & \textbf{28} & \textbf{29} & \textbf{30} & \textcolor{teal}{Avg}\\
\midrule

ERM     & 40.6 $\pm$ 3.8 & 43.1 $\pm$ 4.7 & 39.7 $\pm$ 4.0 & 33.6 $\pm$ 5.0 & 44.7 $\pm$ 6.0 & 31.4 $\pm$ 6.4 & 27.8 $\pm$ 7.7 & 26.9 $\pm$ 6.0 & 27.8 $\pm$ 3.5 & 33.1 $\pm$ 7.1 & 34.9  \\
Mixup   & 40.6 $\pm$ 3.8 & 42.5 $\pm$ 4.7 & 39.7 $\pm$ 4.0 & 33.6 $\pm$ 5.0 & 44.7 $\pm$ 6.0 & 31.4 $\pm$ 6.4 & 27.8 $\pm$ 7.7 & 26.9 $\pm$ 6.0 & 27.8 $\pm$ 3.5 & 33.1 $\pm$ 7.1 & 34.8  \\
MMD     & 38.6 $\pm$ 3.7 & 42.5 $\pm$ 4.7 & 39.7 $\pm$ 4.0 & 33.6 $\pm$ 5.0 & 44.7 $\pm$ 6.0 & 31.4 $\pm$ 6.4 & 27.8 $\pm$ 7.7 & 26.9 $\pm$ 6.0 & 27.8 $\pm$ 3.5 & 33.1 $\pm$ 7.1 & 34.6  \\
MLDG    & 44.4 $\pm$ 4.4 & 43.9 $\pm$ 5.0 & 39.7 $\pm$ 4.0 & 33.6 $\pm$ 5.0 & 44.7 $\pm$ 6.0 & 31.4 $\pm$ 6.4 & 27.8 $\pm$ 7.7 & 26.9 $\pm$ 6.0 & 27.8 $\pm$ 3.5 & 33.1 $\pm$ 7.1 & 35.3  \\
IRM     & 63.6 $\pm$ 4.2 & 56.7 $\pm$ 7.0 & 48.1 $\pm$ 4.3 & 37.2 $\pm$ 4.8 & 44.7 $\pm$ 6.0 & 31.4 $\pm$ 6.4 & 27.8 $\pm$ 7.7 & 26.9 $\pm$ 6.0 & 27.8 $\pm$ 3.5 & 33.1 $\pm$ 7.1 & 39.7  \\
RSC     & 38.6 $\pm$ 3.7 & 42.5 $\pm$ 4.7 & 39.7 $\pm$ 4.0 & 33.6 $\pm$ 5.0 & 44.7 $\pm$ 6.0 & 31.4 $\pm$ 6.4 & 27.8 $\pm$ 7.7 & 26.9 $\pm$ 6.0 & 27.8 $\pm$ 3.5 & 33.1 $\pm$ 7.1 & 34.6  \\
MTL     & 41.7 $\pm$ 5.0 & 45.6 $\pm$ 7.1 & 36.9 $\pm$ 6.3 & 36.4 $\pm$ 6.2 & 44.7 $\pm$ 6.0 & 31.4 $\pm$ 3.2 & 27.8 $\pm$ 4.2 & 28.3 $\pm$ 2.8 & 31.9 $\pm$ 4.6 & 26.1 $\pm$ 4.1 & 35.1  \\
Fish    & 42.5 $\pm$ 3.8 & 43.3 $\pm$ 4.8 & 39.7 $\pm$ 4.0 & 33.6 $\pm$ 5.0 & 44.7 $\pm$ 6.0 & 31.4 $\pm$ 6.4 & 27.8 $\pm$ 7.7 & 26.9 $\pm$ 6.0 & 27.8 $\pm$ 3.5 & 33.1 $\pm$ 7.1 & 35.1  \\
CORAL   & 40.8 $\pm$ 3.7 & 43.3 $\pm$ 4.8 & 39.7 $\pm$ 4.0 & 33.6 $\pm$ 5.0 & 44.7 $\pm$ 6.0 & 31.4 $\pm$ 6.4 & 27.8 $\pm$ 7.7 & 26.9 $\pm$ 6.0 & 27.8 $\pm$ 3.5 & 33.1 $\pm$ 7.1 & 34.9  \\
AndMask & 53.6 $\pm$ 4.2 & 54.7 $\pm$ 5.4 & 51.7 $\pm$ 4.7 & 33.6 $\pm$ 5.1 & 44.7 $\pm$ 6.0 & 31.4 $\pm$ 6.4 & 27.8 $\pm$ 7.7 & 26.9 $\pm$ 6.0 & 27.8 $\pm$ 3.5 & 35.0 $\pm$ 6.7 & 38.9  \\
DIVA    & 40.6 $\pm$ 3.8 & 42.5 $\pm$ 4.7 & 39.7 $\pm$ 4.0 & 33.6 $\pm$ 5.0 & 44.7 $\pm$ 6.0 & 31.4 $\pm$ 6.4 & 27.8 $\pm$ 7.7 & 26.9 $\pm$ 6.0 & 27.8 $\pm$ 3.5 & 33.1 $\pm$ 7.1 & 34.8  \\
LSSAE~(Ours) & 74.5 $\pm$ 1.8 & 65.5 $\pm$ 3.9 & 55.5 $\pm$ 3.2 & 36.0 $\pm$ 3.5 & 45.0 $\pm$ 0.7 & 40.5 $\pm$ 4.6 & 35.0 $\pm$ 1.4 & 33.0 $\pm$ 1.4 & 34.0 $\pm$ 6.4 & 29.0 $\pm$ 2.8 & 44.8  \\
\bottomrule
\end{tabular}
\end{scriptsize}
\end{center}
\vskip -0.1in
\end{table*}


\begin{table*}[!tbp]
\caption{Sine-C. We show the results on each target domain denoted by domain index.}
\vskip 0.15in
\begin{center}
\begin{small}
\begin{tabular}{cccccccccc}
\toprule
Algorithm & \textbf{17} & \textbf{18} & \textbf{19} & \textbf{20} & \textbf{21} & \textbf{22} & \textbf{23} & \textbf{24} & \textcolor{teal}{Avg}\\
\midrule

ERM     & 64.2 $\pm$ 6.8 & 84.9 $\pm$ 3.2 & 83.7 $\pm$ 3.1 & 54.9 $\pm$ 2.1 & 51.1 $\pm$ 6.7 & 54.3 $\pm$ 4.7 & 49.5 $\pm$ 4.8 & 51.7 $\pm$ 5.0 & 61.8  \\
Mixup   & 60.3 $\pm$ 7.8 & 77.4 $\pm$ 3.3 & 87.8 $\pm$ 1.8 & 57.3 $\pm$ 2.6 & 51.1 $\pm$ 6.7 & 54.3 $\pm$ 4.7 & 49.5 $\pm$ 4.8 & 51.7 $\pm$ 5.0 & 61.2  \\
MMD     & 55.8 $\pm$ 5.1 & 57.1 $\pm$ 4.2 & 48.6 $\pm$ 6.4 & 50.7 $\pm$ 1.8 & 51.1 $\pm$ 6.7 & 54.3 $\pm$ 4.7 & 49.5 $\pm$ 4.8 & 51.7 $\pm$ 5.0 & 52.4  \\
MLDG    & 65.6 $\pm$ 6.4 & 88.7 $\pm$ 2.3 & 84.4 $\pm$ 2.8 & 52.4 $\pm$ 2.6 & 51.1 $\pm$ 6.7 & 54.3 $\pm$ 4.7 & 49.5 $\pm$ 4.8 & 51.7 $\pm$ 5.0 & 62.2  \\
IRM     & 61.4 $\pm$ 7.0 & 82.8 $\pm$ 3.2 & 86.5 $\pm$ 3.0 & 54.9 $\pm$ 1.8 & 55.1 $\pm$ 6.7 & 54.3 $\pm$ 4.7 & 49.5 $\pm$ 4.8 & 51.7 $\pm$ 5.0 & 61.5  \\
RSC     & 65.0 $\pm$ 6.7 & 83.5 $\pm$ 3.1 & 85.4 $\pm$ 3.1 & 53.8 $\pm$ 2.6 & 51.1 $\pm$ 6.7 & 54.3 $\pm$ 4.7 & 49.5 $\pm$ 4.8 & 51.7 $\pm$ 5.0 & 61.8  \\
MTL     & 61.9 $\pm$ 7.3 & 82.0 $\pm$ 2.3 & 83.7 $\pm$ 4.2 & 54.2 $\pm$ 5.4 & 60.3 $\pm$ 7.6 & 53.6 $\pm$ 5.2 & 49.5 $\pm$ 5.3 & 46.9 $\pm$ 5.9 & 61.5  \\
Fish    & 69.4 $\pm$ 6.2 & 94.5 $\pm$ 1.5 & 79.5 $\pm$ 3.1 & 52.4 $\pm$ 2.6 & 51.1 $\pm$ 6.7 & 54.3 $\pm$ 4.7 & 49.5 $\pm$ 4.8 & 51.7 $\pm$ 5.0 & 62.8  \\
CORAL   & 70.3 $\pm$ 5.0 & 77.2 $\pm$ 3.6 & 67.0 $\pm$ 4.2 & 52.1 $\pm$ 2.5 & 51.1 $\pm$ 6.7 & 54.3 $\pm$ 4.7 & 49.5 $\pm$ 4.8 & 51.7 $\pm$ 5.0 & 59.2  \\
AndMask & 55.8 $\pm$ 5.1 & 57.1 $\pm$ 4.2 & 48.6 $\pm$ 6.4 & 50.7 $\pm$ 1.8 & 51.1 $\pm$ 6.7 & 54.3 $\pm$ 4.7 & 49.5 $\pm$ 4.8 & 51.7 $\pm$ 5.0 & 52.3  \\
DIVA    & 76.9 $\pm$ 6.1 & 61.1 $\pm$ 5.9 & 47.6 $\pm$ 3.4 & 48.6 $\pm$ 4.2 & 51.1 $\pm$ 7.0 & 52.9 $\pm$ 6.1 & 38.5 $\pm$ 5.0 & 36.5 $\pm$ 6.8 & 51.7  \\
LSSAE~(Ours) & 62.3 $\pm$ 3.1 & 63.2 $\pm$ 6.7 & 57.6 $\pm$ 2.6 & 66.4 $\pm$ 2.6 & 63.5 $\pm$ 3.9 & 59.5 $\pm$ 4.0 & 52.6 $\pm$ 2.2 & 61.3 $\pm$ 1.9 & 60.8  \\
\bottomrule
\end{tabular}
\end{small}
\end{center}
\vskip -0.1in
\end{table*}

\begin{table*}[h]
\caption{PowerSupply. We show the results on each target domain denoted by domain index.}
\vskip 0.15in
\begin{center}
\begin{scriptsize}
\begin{tabular}{cccccccccccc}
\toprule
Algorithm & \textbf{21} & \textbf{22} & \textbf{23} & \textbf{24} & \textbf{25} & \textbf{26} & \textbf{27} & \textbf{28} & \textbf{29} & \textbf{30} & \textcolor{teal}{Avg}\\
\midrule
ERM     & 69.8 $\pm$ 1.4 & 70.0 $\pm$ 1.4 & 69.2 $\pm$ 1.3 & 64.4 $\pm$ 1.5 & 85.8 $\pm$ 1.0 & 76.0 $\pm$ 1.3 & 70.1 $\pm$ 1.5 & 69.8 $\pm$ 1.5 & 69.0 $\pm$ 1.3 & 65.5 $\pm$ 1.5 & 71.0  \\
Mixup   & 69.6 $\pm$ 1.4 & 69.5 $\pm$ 1.5 & 68.3 $\pm$ 1.5 & 64.3 $\pm$ 1.5 & 87.1 $\pm$ 1.0 & 76.6 $\pm$ 1.3 & 70.1 $\pm$ 1.4 & 69.2 $\pm$ 1.3 & 68.1 $\pm$ 1.5 & 65.0 $\pm$ 1.6 & 70.8  \\
MMD     & 70.0 $\pm$ 1.3 & 69.7 $\pm$ 1.4 & 68.7 $\pm$ 1.4 & 64.8 $\pm$ 1.5 & 85.6 $\pm$ 1.0 & 76.1 $\pm$ 1.3 & 70.0 $\pm$ 1.5 & 69.5 $\pm$ 1.4 & 68.7 $\pm$ 1.3 & 65.6 $\pm$ 1.5 & 70.9  \\
MLDG    & 69.7 $\pm$ 1.4 & 69.7 $\pm$ 1.5 & 68.6 $\pm$ 1.5 & 64.6 $\pm$ 1.5 & 86.4 $\pm$ 1.1 & 76.3 $\pm$ 1.4 & 70.1 $\pm$ 1.4 & 69.4 $\pm$ 1.3 & 68.4 $\pm$ 1.5 & 65.6 $\pm$ 1.5 & 70.8  \\
IRM     & 69.8 $\pm$ 1.4 & 69.5 $\pm$ 1.4 & 68.3 $\pm$ 1.4 & 64.1 $\pm$ 1.4 & 87.2 $\pm$ 0.9 & 76.5 $\pm$ 1.3 & 70.0 $\pm$ 1.5 & 69.1 $\pm$ 1.5 & 68.2 $\pm$ 1.3 & 65.0 $\pm$ 1.4 & 70.8  \\
RSC     & 69.9 $\pm$ 1.4 & 69.6 $\pm$ 1.4 & 68.6 $\pm$ 1.4 & 64.4 $\pm$ 1.5 & 86.6 $\pm$ 1.0 & 76.3 $\pm$ 1.3 & 70.0 $\pm$ 1.5 & 69.4 $\pm$ 1.4 & 68.4 $\pm$ 1.3 & 65.4 $\pm$ 1.5 & 70.9  \\
MTL     & 69.6 $\pm$ 1.4 & 69.4 $\pm$ 1.5 & 68.2 $\pm$ 1.6 & 64.2 $\pm$ 1.5 & 87.4 $\pm$ 1.2 & 76.6 $\pm$ 1.3 & 69.9 $\pm$ 1.5 & 69.1 $\pm$ 1.5 & 68.2 $\pm$ 1.5 & 64.6 $\pm$ 1.4 & 70.7  \\
Fish    & 69.7 $\pm$ 1.4 & 69.4 $\pm$ 1.4 & 68.2 $\pm$ 1.4 & 64.2 $\pm$ 1.4 & 87.3 $\pm$ 1.0 & 76.6 $\pm$ 1.3 & 69.9 $\pm$ 1.5 & 69.2 $\pm$ 1.5 & 68.2 $\pm$ 1.3 & 65.2 $\pm$ 1.5 & 70.8  \\
CORAL   & 69.9 $\pm$ 1.4 & 69.7 $\pm$ 1.4 & 68.9 $\pm$ 1.4 & 64.6 $\pm$ 1.4 & 86.1 $\pm$ 1.0 & 76.3 $\pm$ 1.3 & 70.0 $\pm$ 1.5 & 69.5 $\pm$ 1.5 & 68.8 $\pm$ 1.3 & 65.7 $\pm$ 1.5 & 71.0  \\
AndMask & 69.9 $\pm$ 1.4 & 69.4 $\pm$ 1.4 & 68.2 $\pm$ 1.3 & 64.0 $\pm$ 1.4 & 87.4 $\pm$ 0.9 & 76.7 $\pm$ 1.3 & 70.0 $\pm$ 1.5 & 69.1 $\pm$ 1.5 & 68.0 $\pm$ 1.3 & 64.7 $\pm$ 1.5 & 70.7  \\
DIVA    & 69.7 $\pm$ 1.4 & 69.5 $\pm$ 1.3 & 68.2 $\pm$ 1.4 & 63.9 $\pm$ 1.5 & 87.5 $\pm$ 1.0 & 76.5 $\pm$ 1.3 & 69.9 $\pm$ 1.5 & 69.1 $\pm$ 1.5 & 68.1 $\pm$ 1.3 & 64.7 $\pm$ 1.5 & 70.7  \\
LSSAE~(Ours) & 70.0 $\pm$ 1.4 & 69.8 $\pm$ 1.4 & 69.0 $\pm$ 1.5 & 65.4 $\pm$ 1.4 & 85.1 $\pm$ 1.1 & 76.0 $\pm$ 1.4 & 70.1 $\pm$ 1.7 & 69.9 $\pm$ 1.3 & 69.0 $\pm$ 1.6 & 66.3 $\pm$ 1.4 & 71.1  \\
\bottomrule
\end{tabular}
\end{scriptsize}
\end{center}
\vskip -0.1in
\end{table*}

\end{document}